\documentclass{article}

\usepackage[preprint]{neurips_2026}

\usepackage[utf8]{inputenc}
\usepackage[T1]{fontenc}
\usepackage{hyperref}
\usepackage{url}
\usepackage{booktabs}
\usepackage{amsfonts}
\usepackage{amsmath}
\usepackage{amssymb}
\usepackage{nicefrac}
\usepackage{microtype}
\usepackage{xcolor}
\usepackage{graphicx}
\usepackage{wrapfig}
\usepackage{multirow}
\usepackage{makecell}
\usepackage{subcaption}
\usepackage{soul}
\usepackage{enumitem}
\usepackage{placeins}
\usepackage{colortbl}
\usepackage{xspace}
\usepackage{comment}

\newcommand{\eg}{\textit{e.g.}}

\newcommand{\gateprobe}{\textbf{\texttt{GateProbe}}\xspace}
\newcommand{\DTR}{\textbf{DTR}\xspace}
\newcommand{\vd}[2]{#1\textsubscript{\tiny\textcolor{blue!80!black}{#2}}}

\usepackage[most]{tcolorbox}
\newenvironment{findingbox}{%
  \vspace{-8pt}%
  \begin{tcolorbox}[colback=gray!8, colframe=gray!50, boxrule=0.4pt,
    left=4pt, right=4pt, top=2pt, bottom=2pt,
    fontupper=\itshape, before skip=0pt, after skip=0pt]%
}{%
  \end{tcolorbox}\vspace{4pt}%
}

\title{Drop-Then-Recovery: How Redundant Are Vision-Language-Action Models?}

\author{%
  Guoheng Sun$^{1}$\quad Kaixi Feng$^{1}$\quad Shwai He$^{1}$\quad Xiaochuan Gong$^{1}$\\
  \textbf{Yexiao He$^{1}$\quad Ziyao Wang$^{1}$\quad Zheyu Shen$^{1}$\quad Wanghao Ye$^{1}$} \\
  \textbf{Ramana Rao Kompella$^{2}$\quad Gaowen Liu$^{2}$\quad Ang Li$^{1}$}\\[6pt]
  $^{1}$University of Maryland, College Park\quad $^{2}$Cisco Research\\
  \texttt{ghsun@umd.edu}\quad \texttt{angliece@umd.edu}
}

\begin{document}

\maketitle

\begin{abstract}
Vision-Language-Action (VLA) models enable instruction-driven robotic manipulation, but they inherit oversized language backbones from pretrained VLMs whose capacity far exceeds what is needed for short robotic instructions. This raises a basic question: how much of a VLA model is actually necessary for closed-loop control?
In this work, we study architectural redundancy in VLA models by using transformer block removal as a controlled intervention. 
We introduce \textbf{Drop-Then-Recovery (DTR)}, an analysis protocol that removes selected blocks from a pretrained VLA model and then fine-tunes the resulting model to measure whether the removed capacity was necessary for downstream control.
To make this intervention reliable, we propose \textbf{GateProbe}, a one-shot virtual-gate sensitivity metric that ranks blocks by their contribution to the downstream action loss. 
Across multiple VLA architectures, manipulation benchmarks and even real-robot industrial scenarios, we find a strong asymmetry in post-removal recoverability: \ul{\textit{language backbones are highly redundant for standard robotic manipulation tasks, whereas vision and action pathways are substantially less tolerant to removal}}. On LIBERO, removing half of the LLM blocks even improves OpenVLA-OFT from 95.0\% to 98.3\% under the same downstream fine-tuning budget, and retaining only two language blocks still recovers baseline-level performance. These results suggest that current VLA benchmarks may exert limited pressure on deep language grounding and compositional instruction understanding, and that future VLA architectures should allocate capacity more deliberately across language, vision, and action components. 
The code is available at \url{https://github.com/s1ghhh/VLADrop}.

\end{abstract}

\section{Introduction}
\label{sec:intro}

Vision-language-action (VLA) models have become a common framework for instruction-driven robotic manipulation~\citep{brohan2023rt2,kim2024openvla,black2024pi0}. Given visual observations and a natural-language instruction, a VLA policy directly predicts robot actions. Recent systems achieve strong results by combining pretrained vision-language backbones with action prediction modules~\citep{black2025pi05,team2025geminirobotics,kim2025fine}.

A key design choice in these models is to inherit large pretrained backbones. While this provides broad visual and linguistic knowledge~\citep{openxembodiment2023,kim2024openvla,black2024pi0}, it may be excessive for standard manipulation benchmarks, where instructions are often short and templated, such as ``\textit{pick up the red cup}''~\citep{liu2023libero,wang2026vision,fei2025libero,chen2025robotwin2}. This raises a basic question: \textbf{how much of a VLA model is actually needed for closed-loop robotic control?}

This question cannot be answered from parameter count alone. A language block useful for web-scale vision-language modeling may be unnecessary for a downstream control task, while a small action module may be critical because action errors can accumulate over long horizons. Crucially, redundancy must be measured by closed-loop task success after recovery, not merely by single-step prediction loss.

In this work, we study VLA redundancy through \textbf{Drop-Then-Recovery (DTR)}, a controlled protocol that removes transformer blocks and then fine-tunes the remaining model on the downstream task. The goal of DTR is not only to produce a smaller model, but also to measure whether the removed capacity was needed for robotic control: if a dropped model recovers its task success, the removed blocks were not essential for the evaluated task distribution. A central challenge is deciding which blocks to remove: existing layer-dropping metrics often rely on static similarity, parameter magnitude, or immediate degradation~\citep{men2024shortgpt,gromov2024unreasonable,song2024sleb,he2024matters}, but do not necessarily predict \textit{recoverability}. We therefore propose \textbf{GateProbe}, a one-shot virtual-gate metric that ranks blocks by their effect on the downstream action loss.

Our experiments reveal a clear pattern across VLA architectures and manipulation benchmarks: \ul{\textbf{the language backbone is highly redundant under current standard manipulation benchmarks, while the action pathway is much less tolerant to removal}.} 
On LIBERO, dropping half of the LLM blocks already surpasses the full-model baseline under matched training compute, \eg, OpenVLA-OFT reaches 98.3\% vs.\ 95.0\% and $\pi_{0.5}$ reaches 94.0\% vs.\ 91.7\%. Even retaining only two language blocks still matches baseline performance (OpenVLA-OFT 95.1\%; $\pi_{0.5}$ 91.0\%).
In addition to the strong performance of structured Dropping on VLA models, this finding suggests that standard manipulation benchmarks may not fully test language grounding or compositional reasoning, pointing to the need for both more capacity-balanced VLA designs and more linguistically demanding benchmarks.
 Our contributions are summarized:
\begin{itemize}[leftmargin=*,itemsep=0pt,topsep=2pt,parsep=0pt,partopsep=0pt]
    \item \textbf{VLA language backbones are highly over-sized for current manipulation benchmarks.} Across multiple VLA architectures, we find that most language blocks can be removed and recovered with little or no loss in task success.

    \item \textbf{DTR provides a simple way to measure and use this redundancy.} We study redundancy by physically removing transformer blocks from pretrained VLA models and then fine-tuning the smaller models. This protocol tests whether the removed capacity was needed for closed-loop control, while also producing a smaller dense model when the removed blocks are recoverable.

    \item \textbf{GateProbe improves block selection under aggressive removal.}  We propose a one-shot virtual-gate metric that ranks blocks by their effect on the downstream action loss. Compared with static metrics, GateProbe better selects recoverable block sets, especially when only a few language blocks are kept.

    \item \textbf{Current VLA benchmarks may under-test language grounding.}     The ease of recovering from large language-block removal suggests that standard manipulation benchmarks may not require rich language understanding. Our results motivate benchmarks with more compositional instructions, stronger language grounding, and longer-horizon language-conditioned control.
\end{itemize}

\section{Related work}
\label{sec:related}

\textbf{Vision-language-action models.}
VLA models combine pretrained vision-language backbones with action prediction modules for instruction-driven robotic manipulation~\citep{brohan2023rt2,kim2024openvla,black2024pi0}, and recent scaling of both data and backbone capacity has led to strong performance across diverse tasks~\citep{black2025pi05,kim2025fine}.

\textbf{Model compression for LLMs and VLMs.}
Compression techniques for LLMs and VLMs include post-training quantization~\citep{frantar2022gptq}, unstructured or semi-structured pruning~\citep{sun2023simple,frantar2023sparsegpt}, structured pruning~\citep{ma2023llm}, and transformer block removal based on layer-importance criteria~\citep{men2024shortgpt,gromov2024unreasonable,he2025router,song2024sleb,he2024matters}. These methods typically evaluate compressed models \textit{without} recovery fine-tuning. While tolerable for language or vision-language benchmarks, this drop-only regime is insufficient for VLA models, where small action errors compound over long horizons and cause task-level collapse. Our work therefore studies the \textit{compress-then-recover} regime, where post-compression fine-tuning is essential for judging what was truly necessary.

\textbf{Compression and efficiency for VLA models.}
Recent work improves VLA efficiency through quantization, token pruning, layer skipping, distillation and so on~\citep{xu2026qvla,wang2025bitvla,yang2025efficientvla,wang2025specprunevla,zhang2025molevla,chen2025rlrc,jeon2026shallowpi}. Concurrent studies show that naive pruning can harm VLA behavior and that redundancy is asymmetric across components~\citep{jabbour2025scissors,grant2026mechanistic}. These efforts confirm that VLA redundancy exists, but \textit{how redundant} each component is and \textit{where the recoverability limit lies} remain open. 
We use structured block removal as a controlled probe and pair it with \gateprobe, a recoverability-aware importance metric, to systematically evaluate redundancy in VLA models.

\FloatBarrier

\section{Method}
\label{sec:method}

\begin{figure}[htbp]
    \centering
    \includegraphics[width=0.99\textwidth]{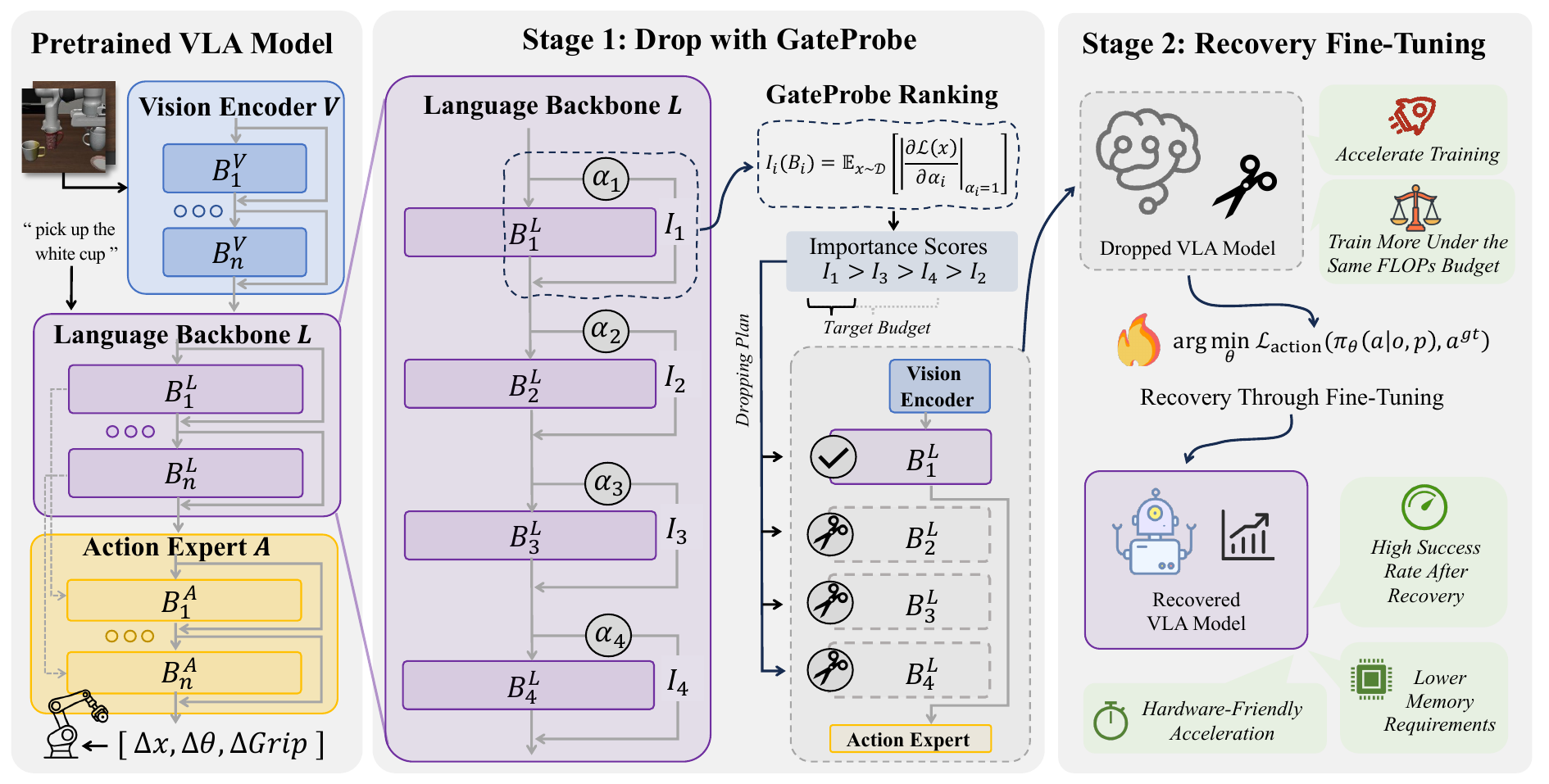} 
    \caption{Overview of DTR. A pretrained VLA model's transformer blocks are ranked by importance, the least important are physically removed, and the smaller model is recovery fine-tuned.}
    \label{fig:overview}
\end{figure}

A VLA model consists of a vision encoder $\mathcal{V}$, a language backbone $\mathcal{L}$, and an action head $\mathcal{A}$, each built from stacked transformer blocks with residual connections. We denote the full set of droppable blocks across all components as
\begin{equation}
    \mathcal{B} = \underbrace{\{B^{\mathcal{V}}_1, \ldots, B^{\mathcal{V}}_{N_V}\}}_{\text{vision}} \;\cup\; \underbrace{\{B^{\mathcal{L}}_1, \ldots, B^{\mathcal{L}}_{N_L}\}}_{\text{language}} \;\cup\; \underbrace{\{B^{\mathcal{A}}_1, \ldots, B^{\mathcal{A}}_{N_A}\}}_{\text{action}},
    \label{eq:block_set}
\end{equation}
where each block $B_i$ follows the residual form $h_i = h_{i-1} + F_i(h_{i-1};\, \theta_i)$.
Note that in dual-stream architectures like $\pi_{0.5}$~\citep{black2025pi05},
dropping a language block does not remove all of its parameters (K/V projections are retained for cross-attention). We detail this mechanism and its impact on compression ratios in Appendix~\ref{app:joint_attention}.

\subsection{The Drop-Then-Recovery (DTR) protocol}
\label{sec:method_dtr}
\label{sec:method_probe}

We use block removal as a structural probe for VLA redundancy. Unlike weight-level compression methods such as quantization or pruning, block removal operates on explicit architectural units while leaving the remaining network as a dense model. This makes the intervention easy to compare across components and less tied to hardware-specific kernels. \DTR formalizes this idea as an analysis protocol that first removes selected transformer blocks and then measures whether the resulting model can recover task performance after fine-tuning. \DTR operates in two stages (Figure~\ref{fig:overview}):

\textbf{Stage 1: Drop.} Given an importance metric $I$ and a target drop count $K$, we select and physically remove the $K$ least important blocks:
\begin{equation}
    \mathcal{S} = \operatorname{argsort}_{K}\bigl\{I(B_i)\bigr\}_{B_i \in \mathcal{B}}, \qquad \mathcal{M}_{\text{drop}} = \mathcal{M} \setminus \mathcal{S}.
    \label{eq:drop}
\end{equation}
Dropping short-circuits the residual ($h_i = h_{i-1}$) and discards $\theta_i$, producing a genuinely smaller dense model with proportionally reduced FLOPs, memory, and latency on any hardware. Since dropping occurs \textit{before} recovery, the smaller model also trains faster and uses less GPU memory.

\textbf{Stage 2: Recovery.} The dropped model is fine-tuned on the downstream task:
\begin{equation}
    \theta^* = \arg\min_{\theta} \; \mathcal{L}_{\text{action}}\!\left(\pi_{\theta}(a \mid o, p),\; a^{\text{gt}}\right),
    \label{eq:recovery}
\end{equation}
where $o$ is the observation, $p$ is the language instruction, $a^{\text{gt}}$ is the demonstration action, and $\mathcal{L}_{\text{action}}$ is the action prediction loss (\eg, MSE for continuous actions, flow-matching for diffusion-based heads). Recovery is critical because in closed-loop control, even small degradations in action quality can compound into task failures over long horizons.

\textbf{Recoverability.} We refer to the task performance of a dropped model \textit{after} recovery fine-tuning as its \textbf{recoverability}. This is distinct from \textit{importance} (how much performance drops immediately upon removal): a block may cause a large zero-shot degradation yet be easily recoverable after fine-tuning.

\subsection{Block selection via \gateprobe}
\label{sec:method_metrics}

A critical component of \DTR is the importance metric $I(B_i)$ used to decide which blocks to drop (Eq.~\ref{eq:drop}). Existing static metrics such as cosine similarity, perplexity, and magnitude (Section~\ref{sec:related}) only measure the immediate performance after dropping~\citep{he2026demystifying}, without capturing a block's \textit{recovery potential} after fine-tuning. Gradient-based methods like Taylor sensitivity can better estimate recoverability, but degrade under extreme compression ratios and are computationally expensive. To address both limitations, we propose \gateprobe.

\textbf{\gateprobe: virtual gate sensitivity.}
\gateprobe is a metric that operates in \textit{activation space} and directly measures the sensitivity of the task loss to each block's \textit{functional contribution}. The key idea is to introduce a virtual scalar gate $\alpha_i$ on each block's residual branch:
\begin{equation}
    \tilde{h}_i = h_{i-1} + \alpha_i \cdot F_i(h_{i-1};\, \theta_i).
    \label{eq:gated_residual}
\end{equation}
Setting $\alpha_i = 0$ is equivalent to dropping block $B_i$; setting $\alpha_i = 1$ recovers the original model. The \gateprobe importance score is the expected absolute sensitivity of the task loss to this gate:
\begin{equation}
    I_{\text{gate}}(B_i) = \mathbb{E}_{x \sim \mathcal{D}} \left[ \left| \frac{\partial \mathcal{L}(x)}{\partial \alpha_i} \right|_{\alpha_i=1} \right].
    \label{eq:gateprobe}
\end{equation}
By the chain rule, this can be computed without explicitly introducing $\alpha_i$ into the model:
\begin{equation}
    \frac{\partial \mathcal{L}}{\partial \alpha_i}\bigg|_{\alpha_i=1} = \left\langle \frac{\partial \mathcal{L}}{\partial h_i},\; F_i(h_{i-1}) \right\rangle,
    \label{eq:gateprobe_chain}
\end{equation}
where $\partial \mathcal{L} / \partial h_i$ is the downstream gradient flowing back through all subsequent layers (available from backpropagation), and $F_i(h_{i-1}) = h_i - h_{i-1}$ is the block's residual contribution (computed from cached hidden states during the forward pass). The score is thus the inner product of the \textit{downstream gradient} and the \textit{block's residual contribution}, capturing both how much the block changes the representation and how much the downstream computation relies on that change. 
Further details, including a Taylor-expansion interpretation and the full algorithm, are provided in Appendix~\ref{app:gateprobe_interp}.

\section{Simulation experiments}
\label{sec:experiments}

\subsection{Setup}
\label{sec:exp_setup}

\textbf{Models.}
We evaluate \DTR on four representative VLA architectures: $\pi_{0.5}$~\citep{black2025pi05}, OpenVLA-OFT~\citep{kim2025fine}, Lingbot-VLA~\citep{wu2026pragmatic}, and GigaBrain-0~\citep{team2025gigabrain}. These models cover different backbone families, model scales, and action-head designs, including continuous regression, flow matching, and diffusion-based action prediction. Detailed architecture descriptions are provided in Appendix~\ref{app:model_details}.

\textbf{Benchmarks.}
We evaluate on three simulation benchmarks: LIBERO~\citep{liu2023libero}, LIBERO-Plus~\citep{fei2025libero}, and RoboTwin 2.0~\citep{chen2025robotwin2}. LIBERO is used for the main redundancy and metric studies, while LIBERO-Plus and RoboTwin 2.0 test robustness and cross-benchmark transfer. Training details and hyperparameters are provided in Appendix~\ref{app:training_details}.

\subsection{Which VLA component is most redundant?}
\label{sec:exp_modules}

\begin{findingbox}
The Language backbone is overwhelmingly more redundant than Vision or Action.
\end{findingbox}

\begin{table}[t]
\centering
\caption{VLA component redundancy on LIBERO (fixed interventions: Drop Half and Keep 2; vision drop lists in Appendix~\ref{app:vision_drop_lists}). $^*$: OpenVLA-OFT action compression reduces hidden dimension (Appendix~\ref{app:openvla_action_drop}).}
\label{tab:module_drop}
\small
\setlength{\tabcolsep}{3.5pt}
\begin{tabular}{l|l|l|cc|cccc|c}
\toprule
\textbf{Model} & \textbf{Setting} & \textbf{Component} & \textbf{Size} & \textbf{FLOPs} & \textbf{Spatial} & \textbf{Object} & \textbf{Goal} & \textbf{Long} & \textbf{Avg.} \\
\midrule
\multirow{7}{*}{\makecell[l]{OpenVLA\\-OFT}}
& Baseline & --- & 100\% & 100\% & 97.2 & 98.4 & 95.6 & 88.6 & 95.0 \\
\cmidrule{2-10}
& \multirow{3}{*}{Drop Half}
& Vision   & 96.4\% & 95.6\% & 82.6 & 99.0 & 77.2 & 76.8 & 83.9 \\
& & Language & 55.5\% & 55.0\% & \textbf{99.0} & \textbf{100.0} & \textbf{97.8} & \textbf{96.4} & \textbf{98.3} \\
& & Action$^*$  & 99.5\% & 100.0\% & 92.4 & 99.0 & 93.6 & 92.8 & 94.5 \\
\cmidrule{2-10}
& \multirow{3}{*}{Keep 2}
& Vision   & 92.2\% & 91.6\% & 80.2 & 97.6 & 92.4 & 50.4 & 80.2 \\
& & Language & 16.6\% & 15.6\% & \textbf{97.2} & 99.0 & \textbf{95.4} & 88.6 & \textbf{95.1} \\
& & Action$^*$  & 99.3\% & 99.9\% & 89.2 & \textbf{99.2} & 76.6 & \textbf{90.8} & 89.0 \\
\midrule
\multirow{7}{*}{$\pi_{0.5}$}
& Baseline & --- & 100\% & 100\% & 96.6 & 95.0 & 93.0 & 82.0 & 91.7 \\
\cmidrule{2-10}
& \multirow{3}{*}{Drop Half}
& Vision   & 89.5\% & 93.3\% & 84.0 & 90.2 & 82.0 & 66.8 & 80.8 \\
& & Language & 60.6\% & 57.9\% & \textbf{98.8} & \textbf{98.6} & 93.4 & 82.4 & \textbf{93.3} \\
& & Action  & 93.8\% & 99.5\% & 95.8 & 95.0 & \textbf{92.4} & \textbf{89.6} & 93.2 \\
\cmidrule{2-10}
& \multirow{3}{*}{Keep 2}
& Vision   & 79.8\% & 87.0\% & 69.4 & 75.2 & 62.4 & 42.6 & 62.4 \\
& & Language & 30.0\% & 25.1\% & \textbf{94.6} & \textbf{96.0} & \textbf{90.6} & \textbf{82.6} & \textbf{91.0} \\
& & Action  & 88.9\% & 99.1\% & 3.6 & 40.8 & 16.0 & 44.4 & 26.2 \\
\bottomrule
\end{tabular}
\vspace{-0.3cm}
\end{table}

To observe VLA redundancy without importance-metric bias, we apply two simple strategies to each component (Vision, Language, Action) independently: Drop Half (removing all odd-indexed blocks) and Keep~2 (retaining only the first and last blocks). Results are shown in Table~\ref{tab:module_drop}.


On both architectures, Language tolerates compression far better than Vision or Action. On OpenVLA-OFT, Language Drop Half removes 44.5\% of parameters while matching or exceeding the baseline SR (98.3\% vs.\ 95.0\%), whereas Vision removes only 3.6\% but drops to 83.9\%. Under extreme compression (Keep~2), Language remains close to baseline, while Vision and Action collapse to 62.4\% and 26.2\% respectively on $\pi_{0.5}$. We thus focus on language block removal in all subsequent experiments.

\subsection{What is the optimal dropping granularity?}
\label{sec:exp_granularity}

\begin{findingbox}
Whole-block dropping is the most cost-effective granularity for compression.
\end{findingbox}

We next investigate the optimal dropping granularity within the language backbone, comparing three options: whole blocks, MHA sublayers only, and MLP sublayers only~\citep{he2024matters,gromov2024unreasonable}.
Table~\ref{tab:granularity} shows the results. 
On OpenVLA-OFT, block dropping (98.3\%) substantially outperforms MHA (91.9\%) and MLP (65.6\%). On $\pi_{0.5}$, all granularities achieve similar SR (93.3\% to 94.1\%), but block dropping compresses the most. 
Therefore, we adopt whole-block dropping as the default for all subsequent experiments.

\begin{table}[t]
\centering
\caption{Effect of dropping granularity on LLM backbone redundancy (Drop Half, LIBERO).}
\label{tab:granularity}
\small
\setlength{\tabcolsep}{4pt}
\begin{tabular}{l|l|cc|cccc|c}
\toprule
\textbf{Model} & \textbf{Target} & \textbf{Size} & \textbf{FLOPs} & \textbf{Spatial} & \textbf{Object} & \textbf{Goal} & \textbf{Long} & \textbf{Avg.} \\
\midrule
\multirow{3}{*}{\makecell[l]{OpenVLA\\-OFT}}
& MHA   & 85.2\% & 84.4\% & 95.0 & 99.2 & \textbf{99.2} & 76.4 & 91.9 \\
& MLP   & 70.3\% & 70.6\% & 83.2 & 96.0 & 15.4 & 72.8 & 65.6 \\
& Block & 55.5\% & 55.0\% & \textbf{99.0} & \textbf{100.0} & 97.8 & \textbf{96.4} & \textbf{98.3} \\
\midrule
\multirow{3}{*}{$\pi_{0.5}$}
& MHA   & 97.0\% & 95.3\% & 96.4 & 98.0 & \textbf{95.8} & 86.2 & \textbf{94.1} \\
& MLP   & 63.7\% & 62.5\% & 96.8 & 97.4 & 93.4 & \textbf{86.8} & 93.6 \\
& Block & 60.6\% & 57.9\% & \textbf{98.8} & \textbf{98.6} & 93.4 & 82.4 & 93.3 \\
\bottomrule
\end{tabular}
\vspace{-0.3cm}
\end{table}

\subsection{Which importance metrics predict recoverability?}
\label{sec:exp_metrics}

\begin{findingbox}
\gateprobe is the most robust block-selection metric, especially under aggressive compression.
\end{findingbox}

\begin{table}[t]
\centering
\caption{Importance metrics for LLM block dropping on $\pi_{0.5}$ / LIBERO (bsz 32, 30K steps). $^\dagger$: These metrics select the same blocks at this drop level. Please see Table~\ref{tab:drop_index} for kept block indices. }
\label{tab:metric_comparison}
\small
\setlength{\tabcolsep}{3.5pt}
\begin{tabular}{l|p{0.4\linewidth}|cccc|c}
\toprule
\textbf{Setting} & \textbf{Metric} & \textbf{Spatial} & \textbf{Object} & \textbf{Goal} & \textbf{Long} & \textbf{Avg.} \\
\midrule

Baseline & --- & 96.6 & 95.0 & 93.0 & 82.0 & 91.7 \\
\midrule
\multirow{8}{*}{\makecell[l]{Drop-9\\(9/18)}}
& Taylor / IGIA$^\dagger$ & 97.0 & 97.0 & \textbf{94.0} & 88.6 & \textbf{94.2} \\
& \gateprobe          & 97.0 & 98.2 & 92.0 & \textbf{88.8} & 94.0 \\
& Fisher              & \textbf{97.2} & 94.4 & 91.0 & 84.4 & 91.8 \\
& Hessian             & 96.0 & 95.6 & 89.6 & 84.0 & 91.3 \\
& CosSim              & 93.8 & \textbf{99.0} & 89.6 & 78.6 & 90.2 \\
& PPL                 & 93.0 & 97.6 & 90.0 & 78.0 & 89.6 \\
& CosSim (contig.)    & 94.6 & 94.8 & 86.4 & 76.4 & 88.0 \\
& Magnitude           & 95.2 & 94.6 & 89.8 & 72.4 & 88.0 \\
\midrule
\multirow{9}{*}{\makecell[l]{Drop-12\\(12/18)}}
& \gateprobe          & \textbf{96.8} & 97.4 & 90.6 & 83.6 & \textbf{92.1} \\
& Taylor              & 95.4 & 97.0 & 90.0 & \textbf{85.8} & 92.0 \\
& IGIA                & \textbf{96.8} & 97.2 & 91.0 & 81.6 & 91.6 \\
& Fisher              & 95.0 & 97.6 & 88.8 & 81.0 & 90.6 \\
& CosSim (contig.)    & 94.6 & 96.4 & \textbf{91.6} & 78.2 & 90.2 \\
& Hessian             & 93.4 & 95.4 & 90.2 & 81.4 & 90.1 \\
& CosSim              & 95.6 & 97.4 & 87.6 & 77.0 & 89.4 \\
& PPL                 & 94.0 & \textbf{97.8} & 90.6 & 71.2 & 88.4 \\
& Magnitude           & 94.8 & 97.0 & 86.0 & 74.2 & 88.0 \\
\midrule
\multirow{6}{*}{\makecell[l]{Drop-16\\(16/18)}}
& \gateprobe / Fisher$^\dagger$ & \textbf{96.6} & 96.8 & \textbf{90.6} & \textbf{84.6} & \textbf{92.2} \\
& Hessian             & 94.6 & 96.2 & 89.6 & 72.8 & 88.3 \\
& PPL                 & 92.2 & 94.2 & 88.8 & 73.6 & 87.2 \\
& Taylor              & 92.6 & \textbf{97.6} & 80.2 & 70.6 & 85.2 \\
& IGIA                & 93.8 & 88.6 & 84.4 & 70.0 & 84.2 \\
& Mag. / CosSim / CosSim (c.)$^\dagger$ & 94.6 & 97.0 & 73.2 & 	62.6 & 81.9 \\
\midrule
\multirow{3}{*}{\makecell[l]{Drop-17\\(17/18)}}
& \gateprobe / Fisher / Hessian / IGIA / PPL$^\dagger$ & \textbf{94.4} & \textbf{97.2} & \textbf{88.2} & \textbf{75.0} & \textbf{88.7} \\
& Taylor              & 91.6 & 94.4 & 79.2 & 72.2 & 84.4 \\
& Mag. / CosSim / CosSim (c.)$^\dagger$ & 92.2 & 90.6 & 82.0 & 70.2 & 83.8 \\
\bottomrule
\end{tabular}
\vspace{-0.3cm}
\end{table}

Table~\ref{tab:metric_comparison} compares metrics on $\pi_{0.5}$ across four drop levels. Non-gradient metrics (CosSim, Magnitude, PPL) are cheap but consistently underperform across all drop levels. Gradient-based metrics (Taylor, IGIA) work well at moderate compression but degrade under extreme settings. \gateprobe achieves the best or second-best at all four levels, with a growing advantage under aggressive compression ($+$3.9 at Drop-16, $+$4.3 at Drop-17). We use \gateprobe as the default in all subsequent experiments. Details of each metric and the kept block indices are provided in Appendix~\ref{app:metrics}.


\begin{figure}[htbp]
\centering
\includegraphics[width=1.03\textwidth]{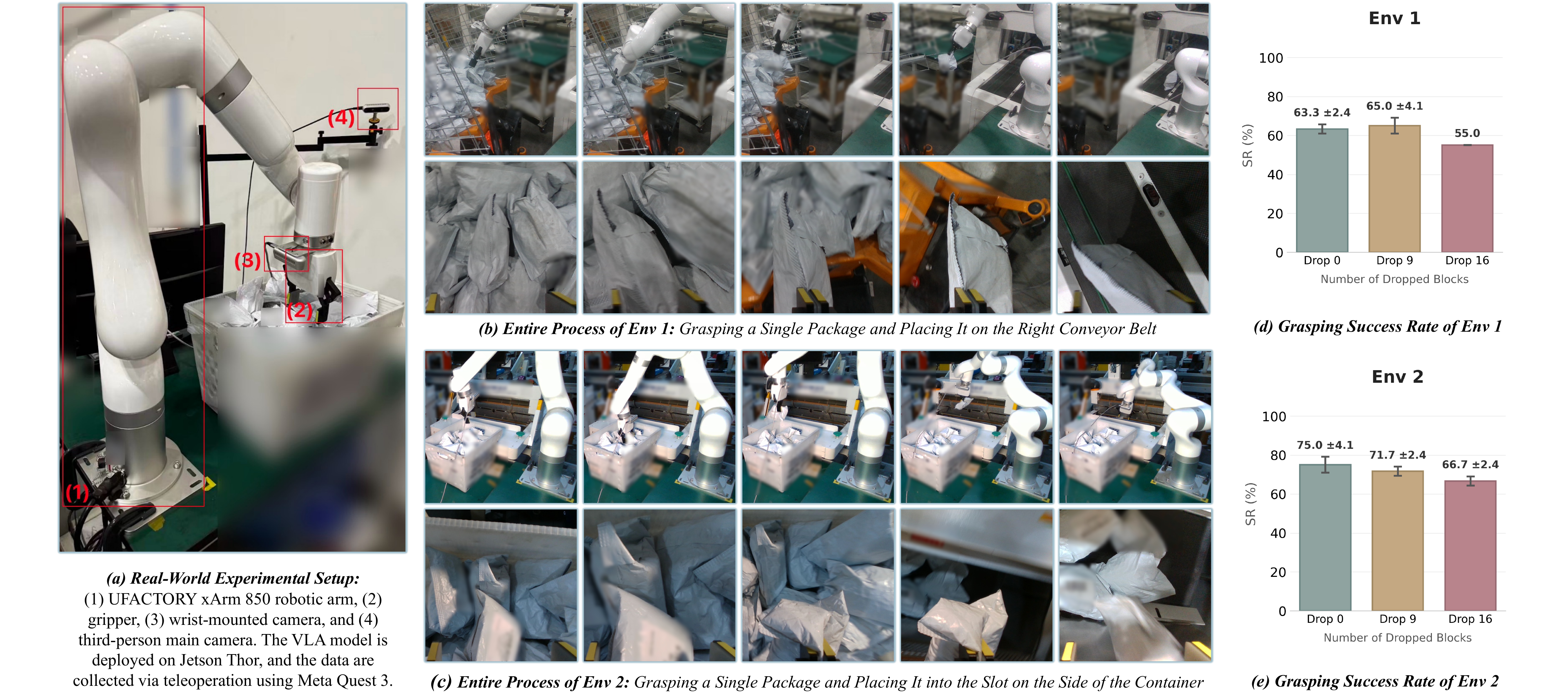}
\caption{Real-world experimental setup and main results. 
}
\label{fig:real_main}
\vspace{-8pt}
\end{figure}

\section{Real-world experiments}
\label{sec:exp_realworld}

\begin{findingbox}
Language-backbone redundancy transfers from simulation to real-world deployment: dropped models perform comparably to the full model in near-training conditions, but exhibit reduced robustness under visual and physical perturbations.
\end{findingbox}

\textbf{Platform and data.}
We deploy $\pi_{0.5}$ on a UFACTORY xArm 850 equipped with a UFACTORY xArm Gripper G2, a wrist-mounted Intel RealSense D435 camera, and a third-person camera (Figure~\ref{fig:real_main}-a). The model runs on an NVIDIA Jetson Thor. Training data are collected via teleoperation with Meta Quest 3 at 10\,Hz, totaling ${\sim}$110K frames (${\sim}$600 grasps). Training follows the setting in Table~\ref{tab:same_flops_libero}.

\textbf{Task.}
The target scenario is warehouse parcel sorting: the robot picks soft-body packages from a filled container and places them onto a conveyor or into adjacent slots. The task poses several challenges: (i)~packages are tightly stacked and deformable, so boundaries are visually ambiguous; (ii)~each package contains a medicine bottle internally, leaving only a limited graspable area on the surface; and (iii)~each attempt must grasp exactly one package. We test two layouts: \textbf{Env\,1} picks from a wire-mesh container to a right-side conveyor (Figure~\ref{fig:real_main}-b), and \textbf{Env\,2} picks from a box to side slots (Figure~\ref{fig:real_main}-c).

\textbf{Main results.}
Figure~\ref{fig:real_main}-d,\,e reports success rates averaged over 3 runs $\times$ 20 grasps. In Env\,1, Drop-9 (65.0\%) slightly exceeds the full model (63.3\%), while Drop-16 degrades to 55.0\%. In Env\,2, the full model achieves 75.0\%, with Drop-9 at 71.7\% and Drop-16 at 66.7\%. This mirrors the simulation pattern: removing half the language blocks preserves task performance, while retaining only 2 of 18 blocks causes moderate degradation.

\begin{figure}[htbp]
\centering
\includegraphics[width=\textwidth]{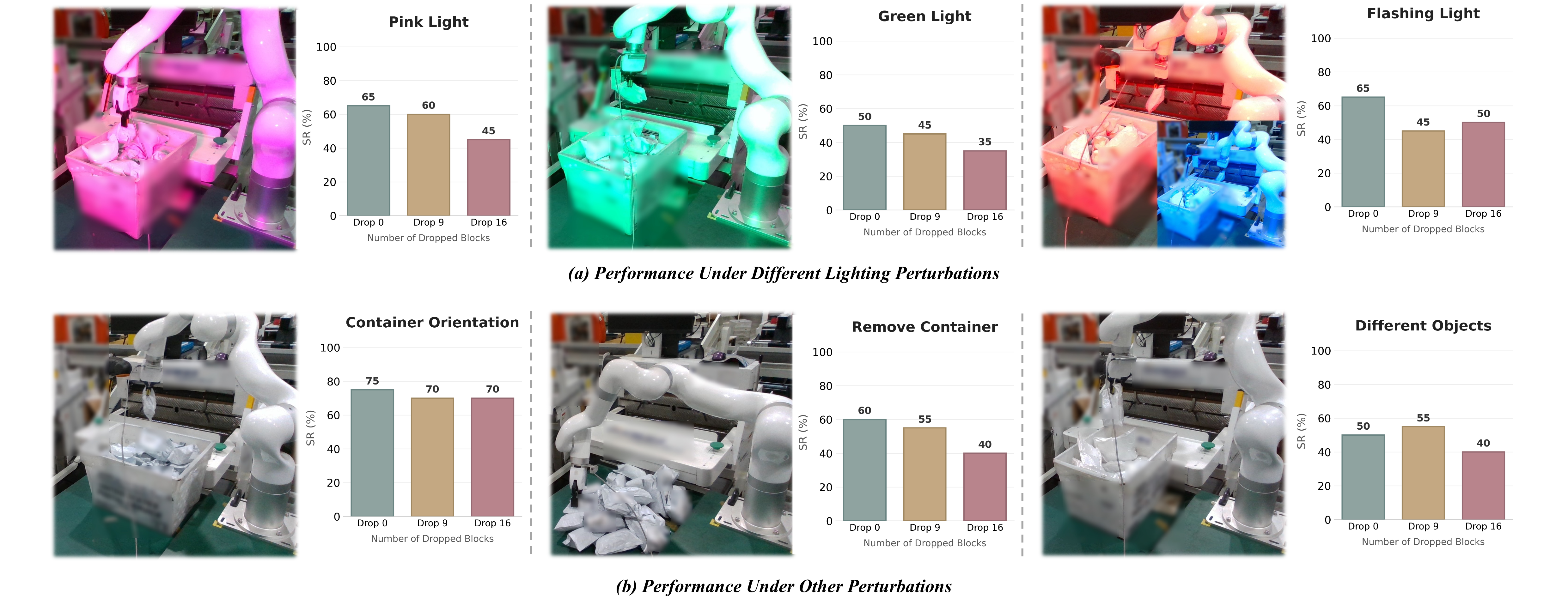}
\caption{Robustness under distribution shift. (a)~Lighting perturbations. (b)~Physical perturbations.}
\label{fig:real_perturb}
\vspace{-0.4cm}
\end{figure}

\textbf{Robustness under distribution shift.}
As shown in Figure~\ref{fig:real_perturb}, we further evaluate under six out-of-distribution (OOD) conditions with a single run of 20 grasps in Env 2, corresponding to Figure~\ref{fig:real_main}-c: three lighting changes (pink, green, flashing), novel objects, altered container orientation, and container removal. 
Under mild perturbations such as container orientation, dropped models remain close to the full model (75\%/70\%/70\% for Drop-0/9/16). However, under stronger perturbations the performance gap becomes evident: for example, under green light Drop-16 falls to 35\% (vs.\ 50\% for the full model), and under container removal Drop-16 drops to 40\% (vs.\ 60\%).
We provide a more detailed analysis of robustness degradation under perturbations in Section~\ref{sec:discussion_benchmark}.

\section{Analysis and discussion}
\label{sec:discussion}

\subsection{What are the practical benefits of reducing redundancy?}
\label{sec:discussion_benefits}

\begin{findingbox}
Removing redundant blocks yields both higher training throughput and hardware-agnostic inference acceleration, orthogonal to existing compression methods.
\end{findingbox}

The redundancy identified above has two direct practical consequences.

\textbf{Higher training throughput under fixed compute.}
Since DTR removes blocks before fine-tuning, a dropped model is cheaper per step, allowing more training iterations under the same compute budget. We verify this by matching total FLOPs to the baseline (bsz~32, 30K steps) through scaled batch size and steps (Table~\ref{tab:same_flops_libero}). Drop-9 through Drop-16 all match or exceed the baseline, with Drop-12 achieving the best average (93.7\%, $+$2.0). Even Drop-17 (a single language block) recovers to 91.0\%.

\begin{table}[t]
\centering
\caption{FLOPs-matched comparison on $\pi_{0.5}$ / LIBERO (baseline: bsz 32 $\times$ 30K steps).}
\label{tab:same_flops_libero}
\small
\setlength{\tabcolsep}{4pt}
\begin{tabular}{l|cc|cc|cccc|c}
\toprule
\textbf{Setting} & \textbf{Size} & \textbf{FLOPs} & \textbf{Bsz} & \textbf{Steps} & \textbf{Spatial} & \textbf{Object} & \textbf{Goal} & \textbf{Long} & \textbf{Avg.} \\
\midrule
Baseline  & 100\% & 100\% & 32 & 30K & 96.6 & 95.0 & 93.0 & 82.0 & 91.7 \\
\midrule
Drop-9    & 60.6\% & 57.9\% & 64 & 25.9K & 92.8 & 97.2 & 91.0 & \textbf{88.2} & 92.3 \\
Drop-12   & 47.5\% & 43.8\% & 64 & 34.2K & \textbf{96.8} & \textbf{98.4} & 94.4 & 85.0 & \textbf{93.7} \\
Drop-16   & 30.0\% & 25.1\% & 64 & 59.8K & 95.2 & 95.6 & \textbf{95.2} & 84.2 & 92.6 \\
Drop-17   & 25.6\% & 20.4\% & 64 & 73.5K & \textbf{96.8} & 95.8 & 92.4 & 78.8 & 91.0 \\
\bottomrule
\end{tabular}
\vspace{-0.4cm}
\end{table}

\textbf{Hardware-agnostic inference acceleration.}
Unlike quantization (which requires low-bit kernel support) and sparse pruning (which requires structured sparsity patterns), DTR produces a standard dense model with fewer layers. The resulting speedup applies to any hardware without specialized kernels or runtime support (see Appendix~\ref{app:edge_acceleration}).

Table~\ref{tab:inference_speedup} benchmarks structural dropping methods on OpenVLA-OFT / LIBERO-Goal. We compare DTR (drop on base model, then fine-tune) with zero-shot CosSim-based dropping (drop blocks directly from the Baseline, no recovery training). We report two speedup measures: \textit{Act. Speedup}, the per-action inference speedup reflecting how fast the model generates a single action; and \textit{Task Speedup} $=$ Act.\ Speedup / Step Ratio, the end-to-end speedup for completing all evaluation episodes. 

DTR-16 achieves \textbf{1.64$\times$} task speedup and also reduces memory by 42\%. Zero-shot attention dropping yields marginal task speedup (Attn Drop 8: 1.09$\times$) due to limited per-step gains. Critically, zero-shot block dropping is \textit{slower} end-to-end despite faster per-action inference: Block Drop 4 achieves 1.05$\times$ action speedup but only 0.72$\times$ task speedup because severe SR degradation (78\%) inflates total steps. This shows that recovery training is not merely beneficial but necessary, without it, the step overhead from degraded SR more than offsets the per-action speedup. 
For more details, please refer to Appendix~\ref{app:compression_comparison}.

\begin{table}[t]
\centering
\caption{Structural dropping methods on OpenVLA-OFT / LIBERO-Goal. Task Speedup $=$ Act.\ Speedup / Step Ratio. We report the latency (ms) and memory (GB) for generating a single action.}
\label{tab:inference_speedup}
\small
\setlength{\tabcolsep}{3pt}
\begin{tabular}{l|c|cc|c|cc}
\toprule
\textbf{Method} & \textbf{SR (\%) $\uparrow$} & \textbf{Act. Speedup $\uparrow$} & \textbf{Latency $\downarrow$} & \textbf{Memory $\downarrow$} & \textbf{Step Ratio $\downarrow$} & \textbf{Task Speedup $\uparrow$} \\
\midrule
Dense Baseline         & 98.0 & 1.00$\times$ & 225.1 & 14.40 & 1.00$\times$ & 1.00$\times$ \\
\midrule
\multicolumn{7}{l}{\textit{Trained block drop (DTR)}} \\
\quad \DTR-16           & \textbf{100.0} & 1.56$\times$ & 144.4 & 8.36 & \textbf{0.95$\times$} & 1.64$\times$ \\
\quad \DTR-30           & 90.0 & \textbf{2.94$\times$} & \textbf{76.7}  & \textbf{3.06}  & 1.13$\times$ & \textbf{2.60$\times$} \\
\midrule
\multicolumn{7}{l}{\textit{CosSim zero-shot drop (applied to fine-tuned baseline, no recovery)}} \\
\quad Attn Drop 4       & \textbf{100.0} & 1.09$\times$ & 206.1 & 13.90 & 0.99$\times$ & 1.10$\times$ \\
\quad Attn Drop 8       & 98.0 & 1.18$\times$ & 191.5 & 13.40 & 1.08$\times$ & 1.09$\times$ \\
\quad Block Drop 4      & 78.0 & 1.05$\times$ & 214.4 & 12.89 & 1.46$\times$ & 0.72$\times$ \\
\quad Block Drop 8      & 18.0 & 1.31$\times$ & 171.5 & 11.38 & 2.39$\times$ & 0.55$\times$ \\
\bottomrule
\end{tabular}
\vspace{-0.4cm}
\end{table}

\subsection{Cross-benchmark analysis: what benchmarks do we need?}
\label{sec:discussion_benchmark}

\begin{figure}[t]
\centering
\includegraphics[width=\textwidth]{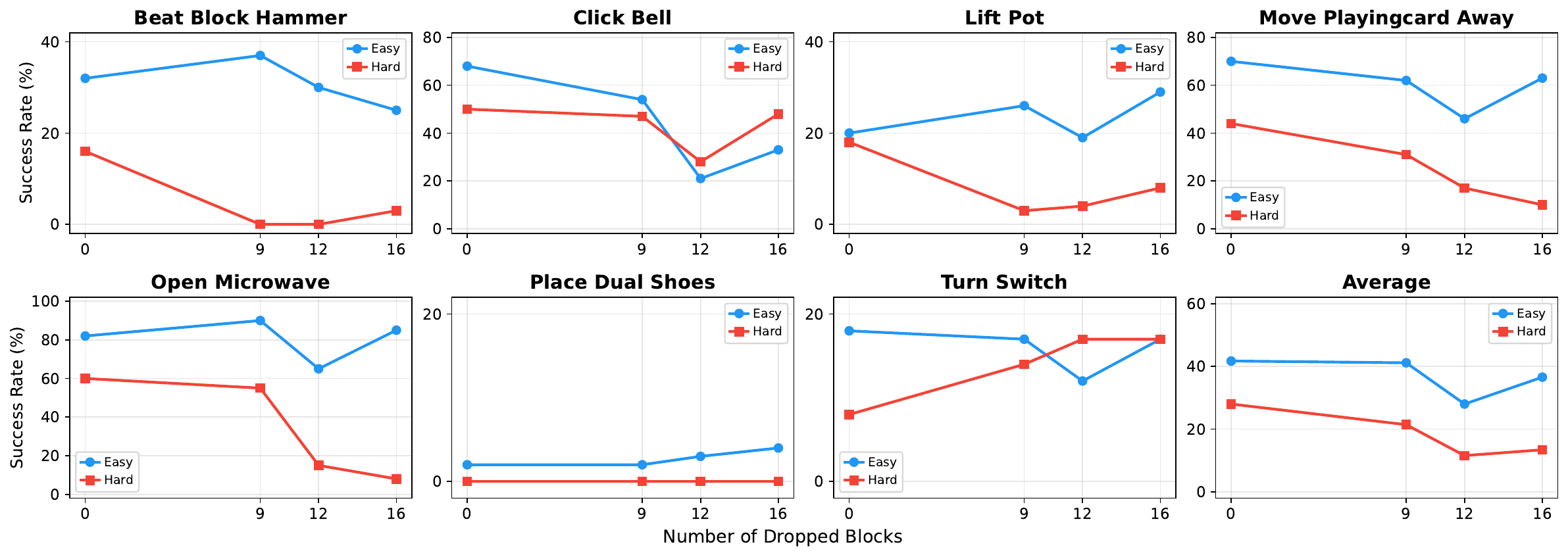}
\caption{Per-task DTR results on RoboTwin 2.0 with $\pi_{0.5}$.}
\label{fig:robotwin_eval}
\vspace{-0.4cm}
\end{figure}

\begin{findingbox}
Language redundancy is universal across benchmarks, but the removed blocks also contribute to physical generalization beyond language understanding.
\end{findingbox}

\begin{table}[t]
\centering
\caption{Per-perturbation-category results on $\pi_{0.5}$ / LIBERO-Plus (bsz 32 for 30K steps). Subscripts indicate relative \% change from baseline. \colorbox{red!40}{Darker red} indicates larger degradation.}
\label{tab:lplus_category}
\small
\setlength{\tabcolsep}{2.5pt}
\begin{tabular}{l|cc|ccccccc|c}
\toprule
\textbf{Setting} & \textbf{Size} & \textbf{FLOPs} & \textbf{Camera} & \textbf{Robot} & \textbf{Language} & \textbf{Light} & \textbf{Background} & \textbf{Noise} & \textbf{Layout} & \textbf{Avg.} \\
\midrule
Baseline & 100\% & 100\% & 85.4 & 60.3 & 70.9 & 91.5 & 92.5 & 87.0 & 82.1 & 81.4 \\
\midrule
Drop-9   & 60.6\% & 57.9\% & \cellcolor{white}\vd{85.4}{-0.0} & \cellcolor{red!22}\vd{49.7}{-10.6} & \cellcolor{red!10}\vd{65.8}{-5.1} & \cellcolor{red!4}\vd{89.7}{-1.8} & \cellcolor{red!6}\vd{89.8}{-2.7} & \cellcolor{red!4}\vd{85.2}{-1.8} & \cellcolor{red!8}\vd{77.6}{-4.5} & \cellcolor{red!8}\vd{77.6}{-3.8} \\
Drop-12  & 47.5\% & 43.8\% & \cellcolor{red!8}\vd{81.8}{-3.6} & \cellcolor{red!36}\vd{43.0}{-17.3} & \cellcolor{red!22}\vd{59.7}{-11.2} & \cellcolor{red!8}\vd{86.9}{-4.6} & \cellcolor{red!10}\vd{87.6}{-4.9} & \cellcolor{red!16}\vd{78.3}{-8.7} & \cellcolor{red!16}\vd{73.6}{-8.5} & \cellcolor{red!16}\vd{73.0}{-8.4} \\
Drop-16  & 30.0\% & 25.1\% & \cellcolor{red!24}\vd{72.7}{-12.7} & \cellcolor{red!50}\vd{36.1}{-24.2} & \cellcolor{red!18}\vd{61.0}{-9.9} & \cellcolor{red!8}\vd{86.8}{-4.7} & \cellcolor{red!18}\vd{82.8}{-9.7} & \cellcolor{red!30}\vd{72.0}{-15.0} & \cellcolor{red!22}\vd{70.4}{-11.7} & \cellcolor{red!24}\vd{68.8}{-12.6} \\
Drop-17  & 25.6\% & 20.4\% & \cellcolor{red!22}\vd{73.7}{-11.7} & \cellcolor{red!50}\vd{32.1}{-28.2} & \cellcolor{red!16}\vd{62.3}{-8.6} & \cellcolor{red!16}\vd{83.3}{-8.2} & \cellcolor{red!24}\vd{79.5}{-13.0} & \cellcolor{red!28}\vd{73.0}{-14.0} & \cellcolor{red!20}\vd{71.8}{-10.3} & \cellcolor{red!26}\vd{68.0}{-13.4} \\
\bottomrule
\end{tabular}
\vspace{-0.4cm}
\end{table}

Following the compute-matched setting from Section~\ref{sec:discussion_benefits}, we evaluate DTR on LIBERO-Plus~\citep{fei2025libero} (Table~\ref{tab:lplus_category}) and RoboTwin 2.0~\citep{chen2025robotwin2} (Figure~\ref{fig:robotwin_eval}; full results in Table~\ref{tab:robotwin_full}).

Language backbone redundancy is consistent across benchmarks, but the magnitude varies. At Drop-9, LIBERO \textit{exceeds} the baseline (92.3\% vs.\ 91.7\%). On LIBERO-Plus, performance drops by 3.8\%. On RoboTwin 2.0, Easy variants decrease by only 0.6\%, while Hard variants degrade by 6.6\%. Furthermore, we make two key observations: (i) on LIBERO-Plus, the largest degradation after dropping is not in the Language category ($-$5.1 at Drop-9) but in Robot ($-$10.6), which perturbs the arm's initial pose. (ii) (ii) Across the seven widely used tasks~\citep{li2025spatial,sun2026rocket} we select from RoboTwin, Hard variants degrade much more sharply than Easy variants across all tasks (Figure~\ref{fig:robotwin_eval}). These suggest that the language backbone, despite being over-parameterized for simple instructions, contributes to the model's \textit{generalization} to physical perturbations. The practical impact of compression thus depends on deployment robustness requirements. 
\ul{\textit{Consequently, we need benchmarks with higher language complexity and sufficiently large out-of-domain perturbations to match the substantial language components in VLA models.}}

\subsection{Cross-model analysis: how should future VLAs be designed?}
\label{sec:discussion_models}

\begin{findingbox}
Language redundancy persists across architectures, rooted in the VLM-to-VLA capacity mismatch.
\end{findingbox}

\begin{wraptable}{r}{0.52\textwidth}
\vspace{-14pt}
\centering
\footnotesize
\caption{\DTR across VLA architectures.}
\label{tab:model_comparison}
\setlength{\tabcolsep}{2pt}
\begin{tabular}{l|l|cccc|c}
\toprule
\textbf{Model} & \textbf{Drop} & \textbf{Spatial} & \textbf{Object} & \textbf{Goal} & \textbf{Long} & \textbf{Avg.} \\
\midrule
\multirow{2}{*}{OpenVLA-OFT}
& 0\,/\,32  & 97.2 & 98.4 & 95.6 & 88.6 & 95.0 \\
& 16\,/\,32 & \textbf{99.0} & \textbf{100.0} & \textbf{97.8} & \textbf{96.4} & \textbf{98.3} \\
\midrule
\multirow{2}{*}{$\pi_{0.5}$}
& 0\,/\,18  & 96.6 & 95.0 & 93.0 & 82.0 & 91.7 \\
& 9\,/\,18  & \textbf{98.8} & \textbf{98.6} & \textbf{93.4} & \textbf{82.4} & \textbf{93.3} \\
\midrule
\multirow{2}{*}{GigaBrain-0}
& 0\,/\,26  & 84.4 & 98.2 & 93.0 & \textbf{76.2} & \textbf{88.0} \\
& 13\,/\,26 & \textbf{85.0} & \textbf{98.6} & \textbf{93.2} & 75.2 & \textbf{88.0} \\
\midrule
\multirow{2}{*}{Lingbot-VLA}
& 0\,/\,36  & 81.8 & 95.0 & \textbf{86.6} & \textbf{67.8} & 82.8 \\
& 18\,/\,36 & \textbf{85.6} & \textbf{97.0} & 84.2 & \textbf{67.8} & \textbf{83.7} \\
\bottomrule
\end{tabular}
\vspace{-2pt}
\end{wraptable}

Table~\ref{tab:model_comparison} follows the setting of Table~\ref{tab:module_drop} and applies the same Drop Half Language protocol to four VLA architectures. Under the same batch size and number of training steps, all four models match or exceed their baselines after removing half of the LLM blocks, with OpenVLA-OFT rebounding by $+3.3$ points. This is partly due to the saturation of LIBERO, and it also confirms that language redundancy is not scale-dependent but instead stems from a structural mismatch: VLA models inherit language capacity far beyond what short robotic instructions require. \ul{\textit{Therefore, for future VLA architecture design, it is necessary to appropriately reduce the language component to better match the task difficulty.}}


\section{Conclusion}
\label{sec:conclusion}
We presented \DTR and \gateprobe to systematically probe architectural redundancy in VLA models. Our findings reveal a striking asymmetry: language backbones inherited from pretrained VLMs carry far more capacity than current robotic manipulation tasks demand, while vision and action pathways remain critical and compress poorly. Both simulation and real-world experiments consistently confirm this pattern across diverse VLA architectures. These results point to a fundamental mismatch between the capacity distribution of today's VLA models and the actual computational demands of closed-loop control, calling for future architectures that allocate capacity more deliberately and future benchmarks that place stronger pressure on compositional language grounding and OOD generalization.

\section*{Acknowledgments}
We would like to thank Dr.\ Hong (Herbert) Cai and Dr.\ Mingu Lee for the many helpful discussions that shaped this work. We also thank the Qualcomm Innovation Fellowship 2025 for generously supporting the authors during this research.


\bibliographystyle{plainnat}
\bibliography{references}

\newpage
\appendix

\section{Model and benchmark details}
\label{app:model_details}

\paragraph{Models.}
We evaluate \DTR on four VLA architectures spanning different backbones, scales, and action-head designs.
(i)~\textbf{$\pi_{0.5}$}~\citep{black2025pi05}: a dual-stream flow-matching architecture with a PaliGemma~\citep{beyer2024paligemma} language backbone (18 Gemma~\citep{team2024gemma} layers) and a separate Gemma action expert (18 layers), using SigLIP~\citep{zhai2023sigmoid} as the vision encoder.
(ii)~\textbf{OpenVLA-OFT}~\citep{kim2025fine}: fine-tunes OpenVLA with a Llama-2-7B~\citep{touvron2023llama} backbone (32 layers), SigLIP vision encoder, and an MLP action head.
(iii)~\textbf{Lingbot-VLA}~\citep{wu2026pragmatic}: built on Qwen2.5-VL-3B~\citep{wang2024qwen2} (36 layers) with a Qwen2 action expert (36 layers) and a flow-matching action decoder, totaling approximately 4B parameters.
(iv)~\textbf{GigaBrain-0}~\citep{team2025gigabrain}: uses PaliGemma2-3B~\citep{steiner2024paligemma} with integrated Gemma2~\citep{team2024gemma2} expert layers (26 layers), SigLIP vision encoder, and a diffusion-based action head, totaling approximately 3.5B parameters.

\paragraph{Benchmarks.}
We use three simulation benchmarks. 
\textbf{LIBERO}~\citep{liu2023libero} contains four task suites: Spatial, Object, Goal, and Long/Libero-10, with 10 tasks per suite. We evaluate each task over 20 trials.
\textbf{LIBERO-Plus}~\citep{fei2025libero} extends LIBERO with increased visual and physical diversity, including perturbations in background, texture, viewpoint, robot pose, language, lighting, noise, and layout.
\textbf{RoboTwin 2.0}~\citep{chen2025robotwin2} targets dual-arm manipulation and is used to evaluate whether language-block redundancy transfers to more complex manipulation settings.

\section{Block dropping in joint-attention architectures}
\label{app:joint_attention}

In $\pi_{0.5}$~\citep{black2025pi05}, the language backbone (PaliGemma) and action expert (Gemma) are parallel transformer streams with identical layer counts. At each layer, both streams perform \textit{joint attention}: the language and action tokens are concatenated along the sequence dimension, and a shared attention matrix is computed over the combined sequence. This means the action expert's queries attend over both its own keys/values and those from the language backbone.

This architecture has a direct implication for block dropping. When we drop a language block, we cannot remove all of its parameters:

(i)~\textbf{Retained parameters.} The language block's \textit{key projection} ($W_K$), \textit{value projection} ($W_V$), and \textit{input layer normalization} must remain active, because the action expert still needs to cross-attend to language representations at this layer.

(ii)~\textbf{Removed parameters.} The language block's \textit{query projection} ($W_Q$), \textit{output projection} ($W_O$), and the entire \textit{MLP sublayer} are eliminated. The language hidden state passes through as identity: $h_i^{\mathcal{L}} = h_{i-1}^{\mathcal{L}}$.

(iii)~\textbf{Gradient flow.} Even after dropping, the retained K/V projections still receive gradients from the action expert's loss during recovery fine-tuning, effectively repurposing them as cross-attention adapters.

As a result, dropping a language block in $\pi_{0.5}$ removes approximately 75\% of that block's parameters (Q, O, and MLP) rather than 100\%. We account for this when computing compression ratios throughout the paper. For architectures without joint attention (\eg, OpenVLA-OFT, where the action head is a separate MLP), dropping a language block removes all of its parameters.

\section{Action head compression in OpenVLA-OFT}
\label{app:openvla_action_drop}

OpenVLA-OFT uses an MLPResNet as its action head, consisting of an input projection (4096 $\to$ $d_h$), two residual MLP blocks ($d_h \to d_h$), and an output projection ($d_h \to 7$), where $d_h$ is the hidden dimension. Unlike the language backbone (which uses block dropping), we compress the action head by reducing $d_h$:
(i)~\textbf{Drop Half}: $d_h = 4096 \to 2048$, reducing action head parameters from 50.4M to 16.8M (model-wide: 99.5\% of original size).
(ii)~\textbf{Extreme}: $d_h = 4096 \to 256$, reducing to 1.3M (model-wide: 99.3\%).
Since the action head accounts for only $\sim$0.7\% of OpenVLA-OFT's total parameters, even aggressive action compression has negligible impact on model size, but can significantly affect task performance (Section~\ref{sec:exp_modules}).

\section{\gateprobe details}
\label{app:gateprobe_interp}
\label{app:gateprobe_algorithm}

\paragraph{Taylor-expansion interpretation.}
$I_{\text{gate}}(B_i)$ is the first-order Taylor approximation of the loss change when the block's contribution is scaled to zero:
\begin{equation}
    \mathcal{L}|_{\alpha_i=0} \approx \mathcal{L}|_{\alpha_i=1} - 1 \cdot \frac{\partial \mathcal{L}}{\partial \alpha_i}\bigg|_{\alpha_i=1}.
    \label{eq:gateprobe_interp}
\end{equation}
A larger $|I_{\text{gate}}(B_i)|$ means removing the block causes a larger immediate loss increase, suggesting it is more important and should be retained.

\paragraph{Implementation.}
The gate in \gateprobe is virtual: we never modify the model architecture or introduce trainable parameters. We register forward pre-hooks at each block's input normalization layer to capture hidden states $h_{i-1}$ and $h_i$, call \texttt{retain\_grad()} on these tensors, and run a standard forward-backward pass. The gate score is then computed as the inner product $\langle \partial \mathcal{L} / \partial h_i, F_i(h_{i-1}) \rangle$ (Eq.~\ref{eq:gateprobe_chain} in the main text). This requires only a single forward-backward pass over a small calibration set.

\begin{figure}[h]
\begin{minipage}{\linewidth}
\small
\begin{center}
\textbf{Algorithm 1:} \gateprobe Importance Profiling
\end{center}
\vspace{-0.5em}
\hrule
\vspace{0.5em}
\textbf{Input:} Model $\mathcal{M}$ with $N$ blocks, calibration data $\{(o_j, p_j, a_j)\}_{j=1}^{M}$, task loss $\mathcal{L}$\\
\textbf{Output:} Importance scores $\{I_{\text{gate}}(B_i)\}_{i=1}^{N}$
\vspace{0.3em}
\hrule
\vspace{0.3em}
\begin{enumerate}[leftmargin=1.5em, itemsep=1pt, topsep=0pt, label=\arabic*.]
    \item Initialize accumulators: $S_i \leftarrow 0$ for $i = 1, \ldots, N$
    \item \textbf{for} each calibration batch $(o, p, a)$ \textbf{do}
    \item \quad Register forward pre-hooks on each block's input norm to capture $h_i$ and call \texttt{retain\_grad()}
    \item \quad Forward pass: compute $\mathcal{L}(o, p, a)$
    \item \quad Backward pass: compute $\partial \mathcal{L} / \partial h_i$ for all $i$
    \item \quad \textbf{for} $i = 1, \ldots, N$ \textbf{do}
    \item \quad\quad $F_i \leftarrow h_i - h_{i-1}$ \hfill $\triangleright$ block residual
    \item \quad\quad $g_i \leftarrow \partial \mathcal{L} / \partial h_i$ \hfill $\triangleright$ downstream gradient
    \item \quad\quad $S_i \leftarrow S_i + |\langle g_i,\, F_i \rangle|$ \hfill $\triangleright$ gate sensitivity
    \item \quad Remove hooks, free intermediate states
    \item $I_{\text{gate}}(B_i) \leftarrow S_i / M$ for all $i$
    \item \textbf{return} $\{I_{\text{gate}}(B_i)\}_{i=1}^{N}$, sorted in descending order
\end{enumerate}
\vspace{0.3em}
\hrule
\label{alg:gateprobe}
\end{minipage}
\end{figure}

\section{Block importance metrics}
\label{app:metrics}

Table~\ref{tab:metric_comparison} compares eight importance metrics for selecting which blocks to drop, most of which originate from prior work on LLM layer pruning. We classify them along two axes: whether they require gradient computation, and whether they operate in parameter space or activation space.

\textbf{Gradient-based, activation-space.}

\textbf{GateProbe} (ours) places a virtual scalar gate $\alpha_l$ (initialized to 1) on the residual branch of each block and measures the sensitivity of the task loss to this gate:
\begin{equation}
  S_l = \mathbb{E}\!\left[\left|\left\langle \frac{\partial \mathcal{L}}{\partial \mathbf{h}_l},\; F_l(\mathbf{h}_{l-1})\right\rangle\right|\right],
\end{equation}
where $F_l(\mathbf{h}_{l-1}) = \mathbf{h}_l - \mathbf{h}_{l-1}$ is the block's residual output and $\mathbf{h}_l$ is the hidden state after block $l$. This is the first-order Taylor approximation of the loss change when scaling the residual by $\alpha_l$. No model modification is needed: hidden states and gradients are captured via hooks. Cost: one forward + backward pass over calibration data.

\textbf{Gradient-based, parameter-space.}

\textbf{Taylor}~\citep{ma2023llm,chen2026prune} accumulates signed first-order salience across calibration batches, then takes the absolute value:
\begin{equation}
  S_l = \frac{1}{N}\sum_{w \in l} \left|\sum_{t=1}^{N} \frac{\partial \mathcal{L}_t}{\partial w} \cdot w\right|.
\end{equation}
Signed accumulation allows positive and negative contributions to cancel, reducing noise. Cost: one forward + backward pass (same as IGIA, but accumulation differs).

\textbf{IGIA}~\citep{huang2026gradpruner} (diagonal Fisher approximation) sums squared gradients per layer across all batches:
\begin{equation}
  S_l = \sum_{t=1}^{N} \sum_{w \in l} \left(\frac{\partial \mathcal{L}_t}{\partial w}\right)^{\!2}.
\end{equation}
This approximates the diagonal of the Fisher Information Matrix. Cost: one forward + backward pass.

\textbf{Fisher}~\citep{molchanov2019importance,chen2026prune} (OBD-style) weights squared gradients by squared parameter magnitudes:
\begin{equation}
  S_l = \frac{1}{N}\sum_{t=1}^{N} \sum_{w \in l} \left(\frac{\partial \mathcal{L}_t}{\partial w}\right)^{\!2} w^2.
\end{equation}
This estimates the loss change from removing each weight under a diagonal Hessian approximation. Cost: one forward + backward pass.

\textbf{Hessian trace}~\citep{chen2026prune} uses Hutchinson's stochastic estimator with Hessian-vector products:
\begin{equation}
  S_l = \left|\operatorname{Tr}(\mathbf{H}_l)\right| \approx \left|\frac{1}{K}\sum_{k=1}^{K} \mathbf{z}_k^\top \mathbf{H}_l \mathbf{z}_k\right|, \quad \mathbf{z}_k \sim \text{Rademacher},
\end{equation}
where $\mathbf{H}_l = \partial^2 \mathcal{L}/\partial \mathbf{w}_l^2$. The Hessian-vector product is computed via double backpropagation without forming $\mathbf{H}_l$ explicitly. Cost: one forward + two backward passes (requires \texttt{create\_graph=True}).

\textbf{Gradient-free, activation-space.}

\textbf{CosSim} (Block Influence) measures how much each block transforms its input:
\begin{equation}
  S_l = 1 - \mathbb{E}\!\left[\cos(\mathbf{h}_{l-1},\, \mathbf{h}_l)\right].
\end{equation}
Blocks with high cosine similarity (low $S_l$) are near-identity and considered more droppable. Proposed by ShortGPT~\citep{men2024shortgpt} and widely adopted as a standard baseline~\citep{he2024matters,chen2026prune}. Cost: one forward pass.

\textbf{CosSim (contig.)} extends this to contiguous block groups. For each possible starting position $s$, it computes the angular distance between $\mathbf{h}_s$ and $\mathbf{h}_{s+n}$ and selects the contiguous window with smallest distance. Follows~\citet{gromov2024unreasonable}. Cost: one forward pass.

\textbf{PPL}~\citep{chen2026prune} (leave-one-out) drops each block individually and measures the resulting loss increase:
\begin{equation}
  S_l = \mathcal{L}(\text{model w/o block } l) - \mathcal{L}(\text{baseline}).
\end{equation}
Cost: $L+1$ forward passes ($L$ = number of layers), making it the most expensive non-gradient method.

\paragraph{Gradient-free, parameter-space.}

\textbf{Magnitude}~\citep{chen2026prune} sums the L1 norm of all parameters in each layer:
\begin{equation}
  S_l = \sum_{w \in l} |w|.
\end{equation}
This is data-independent and deterministic. Cost: zero (no forward pass needed).

\paragraph{Summary.}
Table~\ref{tab:metric_summary} summarizes the classification, computational cost, and wall-clock time measured on $\pi_{0.5}$ (18 PaliGemma blocks) with 64 calibration batches of size 8 on a single H200 GPU.

\begin{table}[h]
\centering
\small
\caption{Classification of block importance metrics. Wall-clock time measured on $\pi_{0.5}$ with 64 calibration batches (batch size 8) on one H200 GPU, excluding model and data loading.}
\label{tab:metric_summary}
\setlength{\tabcolsep}{4pt}
\begin{tabular}{l|cc|l|r}
\toprule
\textbf{Metric} & \textbf{Gradient} & \textbf{Space} & \textbf{Cost} & \textbf{Time (s)} \\
\midrule
\gateprobe (ours) & Yes & Activation & 1 fwd + 1 bwd     &  24.9 \\
IGIA              & Yes & Parameter  & 1 fwd + 1 bwd     &  25.4 \\
Fisher            & Yes & Parameter  & 1 fwd + 1 bwd     &  26.1 \\
Hessian trace     & Yes & Parameter  & 1 fwd + 2 bwd     &  77.2 \\
Taylor            & Yes & Parameter  & 1 fwd + 1 bwd$^*$ & 471.7 \\
\midrule
CosSim            & No  & Activation & 1 fwd              &   9.4 \\
CosSim (contig.)  & No  & Activation & 1 fwd              &   9.2 \\
PPL               & No  & Activation & $(L\!+\!1)$ fwd    & 160.8 \\
Magnitude         & No  & Parameter  & 0                  &   0.2 \\
\bottomrule
\end{tabular}
\vspace{0.3em}
{\\ \footnotesize $^*$ Taylor requires per-parameter signed accumulation on CPU across batches, causing significant memory-transfer overhead despite the same forward/backward count as IGIA and Fisher.}
\end{table}

\section{Drop index lookup tables}
\label{app:drop_index}

Tables~\ref{tab:drop_index},~\ref{tab:drop_index_lplus}, and~\ref{tab:drop_index_robotwin} list the kept PaliGemma block indices for all drop configurations used in this paper. Table~\ref{tab:drop_index} covers the metric comparison on LIBERO (Table~\ref{tab:metric_comparison}); metrics sharing the same row (marked $^\dagger$) select identical blocks at that drop level. Tables~\ref{tab:drop_index_lplus} and~\ref{tab:drop_index_robotwin} cover the \gateprobe selections for LIBERO-Plus (Table~\ref{tab:lplus_category}) and RoboTwin 2.0 (Table~\ref{tab:robotwin_full}), which use dataset-specific profiling and therefore differ from the LIBERO selections.

\begin{table}[h]
\centering
\caption{Kept PaliGemma block indices for each metric and drop configuration on $\pi_{0.5}$ / LIBERO (18 blocks, indexed 0--17). Blocks not listed are dropped.}
\label{tab:drop_index}
\small
\setlength{\tabcolsep}{4pt}
\begin{tabular}{l|l|l}
\toprule
\textbf{Setting} & \textbf{Metric} & \textbf{Keep Blocks} \\
\midrule
\multirow{8}{*}{Drop-9 (9/18)}
& \gateprobe          & [0,1,2,3,4,5,6,8,9] \\
& Taylor / IGIA       & [0,1,3,4,5,6,7,8,9] \\
& Fisher              & [0,1,2,3,4,5,6,7,8] \\
& Hessian             & [0,1,2,3,5,8,9,11,13] \\
& PPL                 & [0,1,4,7,8,10,11,13,14] \\
& CosSim (contig.)    & [0,1,11,12,13,14,15,16,17] \\
& Magnitude           & [2,3,6,12,13,14,15,16,17] \\
& CosSim              & [0,1,2,12,13,14,15,16,17] \\
\midrule
\multirow{9}{*}{Drop-12 (12/18)}
& \gateprobe          & [0,1,2,5,6,8] \\
& Taylor              & [0,3,4,5,6,8] \\
& IGIA                & [0,1,3,4,6,7] \\
& Fisher              & [0,1,5,6,7,8] \\
& Hessian             & [0,1,3,8,9,13] \\
& PPL                 & [0,1,4,7,10,13] \\
& CosSim              & [0,1,14,15,16,17] \\
& Magnitude           & [2,13,14,15,16,17] \\
& CosSim (contig.)    & [0,13,14,15,16,17] \\
\midrule
\multirow{6}{*}{Drop-16 (16/18)}
& \gateprobe / Fisher & [0,5] \\
& Taylor              & [5,6] \\
& IGIA                & [0,7] \\
& Hessian             & [0,1] \\
& PPL                 & [0,13] \\
& Mag. / CosSim / CosSim (c.) & [16,17] \\
\midrule
\multirow{3}{*}{Drop-17 (17/18)}
& \gateprobe / Fisher / Hessian / IGIA / PPL & [0] \\
& Taylor              & [5] \\
& Mag. / CosSim / CosSim (c.) & [17] \\
\bottomrule
\end{tabular}
\end{table}

\begin{table}[h]
\centering
\caption{Kept PaliGemma block indices for \gateprobe on $\pi_{0.5}$ / LIBERO-Plus (18 blocks, indexed 0--17).}
\label{tab:drop_index_lplus}
\small
\setlength{\tabcolsep}{4pt}
\begin{tabular}{l|l}
\toprule
\textbf{Setting} & \textbf{Keep Blocks} \\
\midrule
Drop-9 (9/18)  & [0,1,3,4,5,6,7,8,9] \\
Drop-12 (12/18) & [0,1,4,5,6,7] \\
Drop-16 (16/18) & [0,5] \\
Drop-17 (17/18) & [0] \\
\bottomrule
\end{tabular}
\end{table}

\begin{table}[h]
\centering
\caption{Kept PaliGemma block indices for \gateprobe on $\pi_{0.5}$ / RoboTwin 2.0 (18 blocks, indexed 0--17). Each task uses its own per-task \gateprobe profiling.}
\label{tab:drop_index_robotwin}
\small
\setlength{\tabcolsep}{2.5pt}
\begin{tabular}{l|l|l|l}
\toprule
\textbf{Task} & \textbf{Drop-9} & \textbf{Drop-12} & \textbf{Drop-16} \\
\midrule
Beat Block Hammer   & [0,1,2,3,4,5,6,8,13]  & [0,1,2,3,5,6]   & [0,1] \\
Click Bell          & [0,1,2,3,4,5,6,11,12] & [0,1,2,3,5,11]  & [0,2] \\
Lift Pot            & [0,1,2,3,4,5,6,11,12] & [0,1,2,3,4,5]   & [0,5] \\
Move Playingcard    & [0,1,2,3,4,5,6,7,9]   & [0,1,2,3,5,6]   & [0,2] \\
Open Microwave      & [0,1,2,3,4,5,6,7,13]  & [0,1,2,5,6,7]   & [0,1] \\
Place Dual Shoes    & [0,1,2,3,4,5,6,7,13]  & [0,1,2,3,4,5]   & [0,1] \\
Turn Switch         & [0,1,2,3,4,5,6,9,13]  & [0,1,2,3,5,6]   & [0,2] \\
\bottomrule
\end{tabular}
\end{table}

\section{Vision drop lists}
\label{app:vision_drop_lists}

For all vision-dropping experiments, the attention and MLP drop lists are identical. 
We report the retained vision blocks below. All layer indices are zero-based.

\begin{table}[h]
\centering
\caption{Retained vision block indices for $\pi_{0.5}$ vision-drop experiments. The vision backbone is a SigLIP tower with 27 layers, indexed 0--26.}
\label{tab:vision_drop_list_pi05}
\small
\begin{tabular}{l|l}
\toprule
\textbf{Setting} & \textbf{Retained Blocks} \\
\midrule
Drop Half & $[0,2,4,6,8,10,12,14,16,18,20,22,24,26]$ \\
Keep 2    & $[0,26]$ \\
\bottomrule
\end{tabular}
\end{table}

\begin{table}[h]
\centering
\caption{Retained vision block indices for OpenVLA-OFT vision-drop experiments. OpenVLA-OFT uses a fused DINO+SigLIP vision backbone: DINO has 24 layers indexed 0--23, and SigLIP has 27 layers indexed 0--26.}
\label{tab:vision_drop_list_oft}
\small
\begin{tabular}{l|l|l}
\toprule
\textbf{Setting} & \textbf{DINO Retained Blocks} & \textbf{SigLIP Retained Blocks} \\
\midrule
Drop Half & $[0,2,4,6,8,10,12,14,16,18,20,22]$ 
          & $[0,2,4,6,8,10,12,14,16,18,20,22,24,26]$ \\
Keep 2    & $[0,23]$ 
          & $[0,25]$ \\
\bottomrule
\end{tabular}
\end{table}

For OpenVLA-OFT, the DINO tower follows the OpenVLA feature-extraction path, where the final visual representation is taken from the second-to-last effective block. Therefore, in the Keep 2 setting, layer 23 is retained as the DINO endpoint, while layer 25 is retained as the corresponding SigLIP endpoint.

\section{Training details}
\label{app:training_details}

We use the same recovery fine-tuning protocol as the corresponding full-model baseline whenever possible. This ensures that performance differences mainly reflect the effect of dropping rather than changes in data, optimization, or training budget. For compute-matched comparisons, we scale the training budget according to the compute reduction of the dropped model while keeping the data and optimization protocol unchanged.

\paragraph{LIBERO.}
We train and recover all models on the mixed LIBERO training set, combining the Spatial, Object, Goal, and Long suites from the modified LIBERO RLDS datasets. The main training settings are summarized in Table~\ref{tab:libero_training_details}.

\begin{table}[h]
\centering
\small
\caption{Training settings on LIBERO.}
\label{tab:libero_training_details}
\begin{tabular}{lcc}
\toprule
Setting & OpenVLA-OFT & $\pi_{0.5}$ \\
\midrule
Base checkpoint & OpenVLA-7B & $\pi_{0.5}$ base checkpoint \\
Training data & Spatial, Object, Goal, Long & Spatial, Object, Goal, Long \\
Objective & L1 action regression & Flow-matching action prediction \\
Optimizer & AdamW & AdamW \\
Global batch size & 16 & 32 \\
Training steps & 50K & 30K \\
Learning rate & $5\times10^{-4}$ & $5\times10^{-5}$ \\
LR schedule & decay after 30K steps & 10K warmup, then constant LR \\
Precision & bfloat16 & bfloat16 \\
Fine-tuning & LoRA, rank 32, dropout 0.0 & full fine-tuning \\
Inputs & 2 images + proprioception & LIBERO observation + robot state \\
Image augmentation & enabled & default setting \\
\bottomrule
\end{tabular}
\end{table}

For OpenVLA-OFT, we follow the original fine-tuning recipe with LoRA adaptation, L1 action regression, two input images, proprioceptive inputs, and image augmentation. For $\pi_{0.5}$, we use the standard LIBERO training configuration with a 30K-step budget, global batch size 32, bfloat16 precision, and a learning rate of $5\times10^{-5}$ after a 10K-step warmup.

\paragraph{RoboTwin 2.0.}
We train and recover all models on the mixed 7 RoboTwin 2.0 training set, combining trajectories from the selected RoboTwin 2.0 manipulation tasks. The main training settings are summarized in Table~\ref{tab:RoboTwin 2.0_training_details}.

\begin{table}[h]
\centering
\small
\caption{Training settings on RoboTwin 2.0.}
\label{tab:RoboTwin 2.0_training_details}
\begin{tabular}{lcc}
\toprule
Setting & OpenVLA-OFT & $\pi_{0.5}$ \\
\midrule
Base checkpoint & OpenVLA-7B & $\pi_{0.5}$ base checkpoint \\
Training data & mixed 7 RoboTwin 2.0 tasks & mixed 7 RoboTwin 2.0 tasks \\
Objective & L1 action regression & Flow-matching action prediction \\
Optimizer & AdamW & AdamW \\
Global batch size & 16 & 32 \\
Training steps & 100K & 30K \\
Learning rate & $5\times10^{-4}$ & $5\times10^{-5}$ \\
LR schedule & decay after 30K steps & 10K warmup, then constant LR \\
Precision & bfloat16 & bfloat16 \\
Fine-tuning & LoRA, rank 32, dropout 0.0 & full fine-tuning \\
Inputs & 3 images + proprioception & 3 camera views + robot state \\
Image augmentation & enabled & default setting \\
\bottomrule
\end{tabular}
\end{table}

For OpenVLA-OFT, we use LoRA adaptation with rank 32, L1 action regression, three input images, proprioceptive inputs, and image augmentation. For $\pi_{0.5}$, we use the RoboTwin 2.0 mixed-task training configuration with a 30K-step budget, global batch size 32, bfloat16 precision, and a learning rate of $5\times10^{-5}$ after a 10K-step warmup.

\section{Per-task results on LIBERO-Plus}
\label{app:libero_plus_details}

Table~\ref{tab:lplus_detailed} provides the full per-task results on LIBERO-Plus.

\begin{table}[h]
\centering
\caption{Per-perturbation-category breakdown on LIBERO-Plus for each task suite with \gateprobe block selection. Subscripts indicate absolute difference from baseline.}
\label{tab:lplus_detailed}
\small
\setlength{\tabcolsep}{2.5pt}
\begin{tabular}{l|cc|ccccccc|c}
\toprule
\textbf{Setting} & \textbf{Size} & \textbf{FLOPs} & \textbf{Camera} & \textbf{Robot} & \textbf{Language} & \textbf{Light} & \textbf{Background} & \textbf{Noise} & \textbf{Layout} & \textbf{Avg.} \\
\midrule
\multicolumn{11}{c}{\cellcolor{gray!15}\textit{Spatial}} \\
\midrule
Baseline & 100\% & 100\% & 85.4 & 64.6 & 72.1 & 89.0 & 92.2 & 87.5 & 93.8 & 83.0 \\
Drop-9   & 60.6\% & 57.9\% & \cellcolor{green!6}\vd{85.9}{+0.5} & \cellcolor{red!4}\vd{63.7}{-0.9} & \cellcolor{green!10}\vd{74.6}{+2.5} & \cellcolor{red!4}\vd{87.3}{-1.7} & \cellcolor{red!4}\vd{91.5}{-0.7} & \cellcolor{red!4}\vd{86.0}{-1.5} & \cellcolor{red!8}\vd{90.4}{-3.4} & \cellcolor{red!4}\vd{82.4}{-0.6} \\
Drop-12  & 47.5\% & 43.8\% & \cellcolor{red!6}\vd{83.2}{-2.2} & \cellcolor{red!28}\vd{50.6}{-14.0} & \cellcolor{red!4}\vd{70.3}{-1.8} & \cellcolor{red!6}\vd{86.3}{-2.7} & \cellcolor{red!8}\vd{89.1}{-3.1} & \cellcolor{red!16}\vd{80.6}{-6.9} & \cellcolor{red!14}\vd{87.3}{-6.5} & \cellcolor{red!10}\vd{77.6}{-5.4} \\
Drop-16  & 30.0\% & 25.1\% & \cellcolor{red!8}\vd{81.1}{-4.3} & \cellcolor{red!20}\vd{53.7}{-10.9} & \cellcolor{green!6}\vd{72.3}{+0.2} & \cellcolor{green!6}\vd{90.4}{+1.4} & \cellcolor{red!4}\vd{91.5}{-0.7} & \cellcolor{red!10}\vd{82.3}{-5.2} & \cellcolor{red!16}\vd{85.5}{-8.3} & \cellcolor{red!8}\vd{78.8}{-4.2} \\
Drop-17  & 25.6\% & 20.4\% & \cellcolor{green!6}\vd{86.7}{+1.3} & \cellcolor{red!34}\vd{46.6}{-18.0} & \cellcolor{green!10}\vd{74.1}{+2.0} & \cellcolor{green!10}\vd{91.8}{+2.8} & \cellcolor{green!10}\vd{94.6}{+2.4} & \cellcolor{green!6}\vd{87.7}{+0.2} & \cellcolor{red!8}\vd{90.1}{-3.7} & \cellcolor{red!6}\vd{81.0}{-2.0} \\
\midrule
\multicolumn{11}{c}{\cellcolor{gray!15}\textit{Object}} \\
\midrule
Baseline & 100\% & 100\% & 95.7 & 56.5 & 83.6 & 99.0 & 98.8 & 96.2 & 92.3 & 88.1 \\
Drop-9   & 60.6\% & 57.9\% & \cellcolor{green!6}\vd{96.5}{+0.8} & \cellcolor{red!42}\vd{34.2}{-22.3} & \cellcolor{red!16}\vd{76.0}{-7.6} & \cellcolor{red!4}\vd{98.7}{-0.3} & \cellcolor{red!4}\vd{98.4}{-0.4} & \cellcolor{green!6}\vd{97.2}{+1.0} & \cellcolor{red!8}\vd{89.1}{-3.2} & \cellcolor{red!10}\vd{83.1}{-5.0} \\
Drop-12  & 47.5\% & 43.8\% & \cellcolor{red!4}\vd{95.2}{-0.5} & \cellcolor{red!44}\vd{32.2}{-24.3} & \cellcolor{red!14}\vd{76.8}{-6.8} & \cellcolor{red!6}\vd{97.0}{-2.0} & \cellcolor{red!8}\vd{95.6}{-3.2} & \cellcolor{red!8}\vd{92.2}{-4.0} & \cellcolor{red!16}\vd{84.4}{-7.9} & \cellcolor{red!16}\vd{80.7}{-7.4} \\
Drop-16  & 30.0\% & 25.1\% & \cellcolor{red!8}\vd{91.9}{-3.8} & \cellcolor{red!46}\vd{29.9}{-26.6} & \cellcolor{red!8}\vd{79.4}{-4.2} & \cellcolor{red!4}\vd{97.6}{-1.4} & \cellcolor{red!10}\vd{93.5}{-5.3} & \cellcolor{white}\vd{96.2}{0.0} & \cellcolor{red!16}\vd{84.4}{-7.9} & \cellcolor{red!16}\vd{80.7}{-7.4} \\
Drop-17  & 25.6\% & 20.4\% & \cellcolor{red!6}\vd{93.2}{-2.5} & \cellcolor{red!48}\vd{27.6}{-28.9} & \cellcolor{red!6}\vd{81.6}{-2.0} & \cellcolor{red!8}\vd{94.3}{-4.7} & \cellcolor{red!16}\vd{89.9}{-8.9} & \cellcolor{red!6}\vd{94.1}{-2.1} & \cellcolor{red!18}\vd{82.9}{-9.4} & \cellcolor{red!16}\vd{79.5}{-8.6} \\
\midrule
\multicolumn{11}{c}{\cellcolor{gray!15}\textit{Goal}} \\
\midrule
Baseline & 100\% & 100\% & 79.9 & 61.6 & 62.9 & 94.3 & 93.6 & 83.1 & 61.4 & 74.8 \\
Drop-9   & 60.6\% & 57.9\% & \cellcolor{green!18}\vd{88.2}{+8.3} & \cellcolor{red!8}\vd{57.2}{-4.4} & \cellcolor{red!18}\vd{53.4}{-9.5} & \cellcolor{red!4}\vd{93.5}{-0.8} & \cellcolor{green!6}\vd{95.0}{+1.4} & \cellcolor{green!10}\vd{85.8}{+2.7} & \cellcolor{green!6}\vd{61.6}{+0.2} & \cellcolor{red!4}\vd{74.4}{-0.4} \\
Drop-12  & 47.5\% & 43.8\% & \cellcolor{red!4}\vd{79.4}{-0.5} & \cellcolor{red!30}\vd{45.7}{-15.9} & \cellcolor{red!34}\vd{44.1}{-18.8} & \cellcolor{red!14}\vd{88.2}{-6.1} & \cellcolor{red!10}\vd{88.6}{-5.0} & \cellcolor{red!16}\vd{75.5}{-7.6} & \cellcolor{red!8}\vd{57.4}{-4.0} & \cellcolor{red!16}\vd{66.3}{-8.5} \\
Drop-16  & 30.0\% & 25.1\% & \cellcolor{red!50}\vd{48.3}{-31.6} & \cellcolor{red!50}\vd{19.8}{-41.8} & \cellcolor{red!50}\vd{30.7}{-32.2} & \cellcolor{red!18}\vd{84.9}{-9.4} & \cellcolor{red!36}\vd{74.0}{-19.6} & \cellcolor{red!50}\vd{42.7}{-40.4} & \cellcolor{red!22}\vd{49.9}{-11.5} & \cellcolor{red!48}\vd{47.2}{-27.6} \\
Drop-17  & 25.6\% & 20.4\% & \cellcolor{red!42}\vd{56.6}{-23.3} & \cellcolor{red!50}\vd{20.5}{-41.1} & \cellcolor{red!42}\vd{39.3}{-23.6} & \cellcolor{red!30}\vd{79.2}{-15.1} & \cellcolor{red!40}\vd{71.9}{-21.7} & \cellcolor{red!48}\vd{53.3}{-29.8} & \cellcolor{red!10}\vd{55.8}{-5.6} & \cellcolor{red!42}\vd{51.6}{-23.2} \\
\midrule
\multicolumn{11}{c}{\cellcolor{gray!15}\textit{Long}} \\
\midrule
Baseline & 100\% & 100\% & 80.9 & 58.8 & 66.3 & 83.2 & 86.2 & 81.3 & 82.7 & 76.4 \\
Drop-9   & 60.6\% & 57.9\% & \cellcolor{red!18}\vd{71.8}{-9.1} & \cellcolor{red!26}\vd{45.0}{-13.8} & \cellcolor{red!10}\vd{60.6}{-5.7} & \cellcolor{red!8}\vd{78.5}{-4.7} & \cellcolor{red!20}\vd{75.8}{-10.4} & \cellcolor{red!16}\vd{72.8}{-8.5} & \cellcolor{red!28}\vd{68.6}{-14.1} & \cellcolor{red!18}\vd{66.9}{-9.5} \\
Drop-12  & 47.5\% & 43.8\% & \cellcolor{red!20}\vd{70.2}{-10.7} & \cellcolor{red!28}\vd{44.5}{-14.3} & \cellcolor{red!32}\vd{49.6}{-16.7} & \cellcolor{red!16}\vd{75.2}{-8.0} & \cellcolor{red!16}\vd{78.5}{-7.7} & \cellcolor{red!30}\vd{65.9}{-15.4} & \cellcolor{red!34}\vd{64.7}{-18.0} & \cellcolor{red!26}\vd{63.1}{-13.3} \\
Drop-16  & 30.0\% & 25.1\% & \cellcolor{red!20}\vd{70.6}{-10.3} & \cellcolor{red!30}\vd{43.8}{-15.0} & \cellcolor{red!4}\vd{64.8}{-1.5} & \cellcolor{red!20}\vd{73.0}{-10.2} & \cellcolor{red!22}\vd{74.4}{-11.8} & \cellcolor{red!30}\vd{65.7}{-15.6} & \cellcolor{red!40}\vd{61.5}{-21.2} & \cellcolor{red!24}\vd{64.2}{-12.2} \\
Drop-17  & 25.6\% & 20.4\% & \cellcolor{red!38}\vd{60.4}{-20.5} & \cellcolor{red!42}\vd{35.9}{-22.9} & \cellcolor{red!18}\vd{56.9}{-9.4} & \cellcolor{red!32}\vd{66.4}{-16.8} & \cellcolor{red!40}\vd{64.4}{-21.8} & \cellcolor{red!42}\vd{58.1}{-23.2} & \cellcolor{red!46}\vd{56.7}{-26.0} & \cellcolor{red!38}\vd{56.3}{-20.1} \\
\midrule
\multicolumn{11}{c}{\cellcolor{gray!15}\textit{Average (all 4 suites)}} \\
\midrule
Baseline & 100\% & 100\% & 85.4 & 60.3 & 70.9 & 91.5 & 92.5 & 87.0 & 82.1 & 81.4 \\
Drop-9   & 60.6\% & 57.9\% & \cellcolor{white}\vd{85.4}{0.0} & \cellcolor{red!20}\vd{49.7}{-10.6} & \cellcolor{red!10}\vd{65.8}{-5.1} & \cellcolor{red!4}\vd{89.7}{-1.8} & \cellcolor{red!6}\vd{89.8}{-2.7} & \cellcolor{red!4}\vd{85.2}{-1.8} & \cellcolor{red!8}\vd{77.6}{-4.5} & \cellcolor{red!8}\vd{77.6}{-3.8} \\
Drop-12  & 47.5\% & 43.8\% & \cellcolor{red!8}\vd{81.8}{-3.6} & \cellcolor{red!32}\vd{43.0}{-17.3} & \cellcolor{red!22}\vd{59.7}{-11.2} & \cellcolor{red!8}\vd{86.9}{-4.6} & \cellcolor{red!10}\vd{87.6}{-4.9} & \cellcolor{red!16}\vd{78.3}{-8.7} & \cellcolor{red!16}\vd{73.6}{-8.5} & \cellcolor{red!16}\vd{73.0}{-8.4} \\
Drop-16  & 30.0\% & 25.1\% & \cellcolor{red!24}\vd{72.7}{-12.7} & \cellcolor{red!44}\vd{36.1}{-24.2} & \cellcolor{red!18}\vd{61.0}{-9.9} & \cellcolor{red!8}\vd{86.8}{-4.7} & \cellcolor{red!18}\vd{82.8}{-9.7} & \cellcolor{red!30}\vd{72.0}{-15.0} & \cellcolor{red!22}\vd{70.4}{-11.7} & \cellcolor{red!24}\vd{68.8}{-12.6} \\
Drop-17  & 25.6\% & 20.4\% & \cellcolor{red!22}\vd{73.7}{-11.7} & \cellcolor{red!48}\vd{32.1}{-28.2} & \cellcolor{red!16}\vd{62.3}{-8.6} & \cellcolor{red!16}\vd{83.3}{-8.2} & \cellcolor{red!24}\vd{79.5}{-13.0} & \cellcolor{red!28}\vd{73.0}{-14.0} & \cellcolor{red!20}\vd{71.8}{-10.3} & \cellcolor{red!26}\vd{68.0}{-13.4} \\
\bottomrule
\end{tabular}
\end{table}

\section{Full per-task results on RoboTwin 2.0}
\label{app:robotwin_details}

Table~\ref{tab:robotwin_full} provides the full per-task numerical results corresponding to Figure~\ref{fig:robotwin_eval}. We report success rates (\%) for Easy (clean) and Hard (randomized) evaluation variants on each of the 7 RoboTwin 2.0 tasks under different drop levels.

\begin{table}[h]
\centering
\caption{Full per-task results on RoboTwin 2.0 with $\pi_{0.5}$ (\gateprobe block selection). Success rate (\%) for Easy and Hard variants across drop levels. Avg is the mean of Easy and Hard.}
\label{tab:robotwin_full}
\small
\setlength{\tabcolsep}{2.5pt}
\begin{tabular}{l|ccc|ccc|ccc|ccc}
\toprule
& \multicolumn{3}{c|}{\textbf{Baseline}} & \multicolumn{3}{c|}{\textbf{Drop-9}} & \multicolumn{3}{c|}{\textbf{Drop-12}} & \multicolumn{3}{c}{\textbf{Drop-16}} \\
\textbf{Task} & Easy & Hard & Avg & Easy & Hard & Avg & Easy & Hard & Avg & Easy & Hard & Avg \\
\midrule
Beat Block Hammer   & 32 & 16 & 24.0 & 37 &  0 & 18.5 & 30 &  0 & 15.0 & 25 &  3 & 14.0 \\
Click Bell          & 68 & 50 & 59.0 & 54 & 47 & 50.5 & 21 & 28 & 24.5 & 33 & 48 & 40.5 \\
Lift Pot            & 20 & 18 & 19.0 & 26 &  3 & 14.5 & 19 &  4 & 11.5 & 29 &  8 & 18.5 \\
Move Playingcard    & 70 & 44 & 57.0 & 62 & 31 & 46.5 & 46 & 17 & 31.5 & 63 & 10 & 36.5 \\
Open Microwave      & 82 & 60 & 71.0 & 90 & 55 & 72.5 & 65 & 15 & 40.0 & 85 &  8 & 46.5 \\
Place Dual Shoes    &  2 &  0 &  1.0 &  2 &  0 &  1.0 &  3 &  0 &  1.5 &  4 &  0 &  2.0 \\
Turn Switch         & 18 &  8 & 13.0 & 17 & 14 & 15.5 & 12 & 17 & 14.5 & 17 & 17 & 17.0 \\
\midrule
\textbf{Average}    & 41.7 & 28.0 & 34.9 & 41.1 & 21.4 & 31.3 & 28.0 & 11.6 & 19.8 & 36.6 & 13.4 & 25.0 \\
\bottomrule
\end{tabular}
\end{table}

\section{Full compression comparison on LIBERO-Goal}
\label{app:compression_comparison}

Here we report the full 12-method comparison on OpenVLA-OFT / LIBERO-Goal (10 tasks $\times$ 5 trials = 50 episodes, max 300 steps, seed 7), including INT4 quantization and Wanda 2:4 pruning~\citep{sun2023simple} that are omitted from the main text.

\paragraph{Setup.}
The Dense Baseline and DTR results are taken from the OpenVLA-OFT rows in Table~\ref{tab:module_drop}. For CosSim-based zero-shot dropping, we profile the baseline on LIBERO 4-suite calibration data (64 batches $\times$ 8 = 512 samples) to compute Block Influence (BI) $= 1 - \cos(\mathbf{h}_\text{in}, \mathbf{h}_\text{out})$ at block, attention, and MLP granularity. Layers with lowest BI (closest to identity) are dropped first. Wanda 2:4 checkpoints are created from baseline activation norms. Latency and memory are measured on a single H200 GPU using the inference pipeline (VLA forward + action head prediction, batch size 1, 32 iterations, 5-iteration warmup, \texttt{torch.cuda.synchronize}).

\begin{table}[h]
\centering
\caption{Full comparison: 12 compression methods on OpenVLA-OFT / LIBERO-Goal. ``Step Ratio'' compares total environment steps (including failed episodes at 300 max steps) against the baseline. Task Speedup $=$ Act.\ Speedup / Step Ratio.}
\label{tab:full_compression}
\small
\setlength{\tabcolsep}{3pt}
\begin{tabular}{l|c|cc|c|cc}
\toprule
\textbf{Method} & \textbf{SR (\%) $\uparrow$} & \textbf{Act. Speedup $\uparrow$} & \textbf{Latency $\downarrow$} & \textbf{Memory $\downarrow$} & \textbf{Step Ratio $\downarrow$} & \textbf{Task Speedup $\uparrow$} \\
\midrule
Dense Baseline              & 98.0 & 1.00$\times$ & 225.1 & 14.40 & 1.00$\times$ & 1.00$\times$ \\
\midrule
\multicolumn{7}{l}{\textit{Trained block drop}} \\
\quad DTR-16                & \textbf{100.0} & 1.56$\times$ & 144.4 &  8.36 & \textbf{0.95$\times$} & 1.64$\times$ \\
\quad DTR-30                &  90.0 & \textbf{2.94$\times$} & \textbf{76.7}  & \textbf{3.06}  & 1.13$\times$ & \textbf{2.60$\times$} \\
\midrule
\multicolumn{7}{l}{\textit{Traditional compression (applied to baseline, no retraining)}} \\
\quad INT4 Quantization     &  94.0 & 0.61$\times$ & 369.5 &  4.71 & 1.10$\times$ & 0.55$\times$ \\
\quad Wanda 2:4 (LLM only) &  42.0 & 0.99$\times$ & 226.8 & 14.40 & 1.95$\times$ & 0.51$\times$ \\
\quad Wanda 2:4 (full)      &  28.0 & 1.03$\times$ & 217.5 & 14.40 & 2.24$\times$ & 0.46$\times$ \\
\midrule
\multicolumn{7}{l}{\textit{CosSim zero-shot drop (applied to baseline, no retraining)}} \\
\quad Block Drop 4          &  78.0 & 1.05$\times$ & 214.4 & 12.89 & 1.46$\times$ & 0.72$\times$ \\
\quad Block Drop 8          &  18.0 & 1.31$\times$ & 171.5 & 11.38 & 2.39$\times$ & 0.55$\times$ \\
\quad Attn Drop 4           & \textbf{100.0} & 1.09$\times$ & 206.1 & 13.90 & 0.99$\times$ & 1.10$\times$ \\
\quad Attn Drop 8           &  98.0 & 1.18$\times$ & 191.5 & 13.40 & 1.08$\times$ & 1.09$\times$ \\
\quad MLP Drop 4            &   0.0 & 1.15$\times$ & 196.1 & 13.40 & 2.71$\times$ & 0.42$\times$ \\
\quad MLP Drop 8            &   0.0 & 1.19$\times$ & 188.6 & 12.39 & 2.71$\times$ & 0.44$\times$ \\
\bottomrule
\end{tabular}
\end{table}

\paragraph{Key observations.}

\textit{DTR is the only method that improves all three axes.} DTR-16 achieves 100\% SR ($+$2.0 over baseline), 1.64$\times$ task speedup, and 42\% memory savings. No other method improves even two axes simultaneously.

\textit{Per-action speedup does not imply end-to-end speedup.} We report two speedup measures: \textit{Act.\ Speedup}, the per-action inference speedup; and \textit{Task Speedup} $=$ Act.\ Speedup / Step Ratio, the end-to-end speedup for completing all evaluation episodes. Methods that degrade SR inflate total environment steps (failed episodes run to the 300-step horizon), so a faster per-action model can be \textit{slower} overall. For example, Block Drop 4 achieves 1.05$\times$ action speedup but only 0.72$\times$ task speedup, and Wanda 2:4 (full) achieves 1.03$\times$ action speedup but only 0.46$\times$ task speedup. In contrast, DTR-16 combines 1.56$\times$ action speedup with a favorable step ratio (0.95$\times$), yielding 1.64$\times$ task speedup. This shows that recovery training is not merely beneficial but necessary: without it, step overhead from degraded SR more than offsets per-action gains.

\textit{Traditional compression underperforms in this kernel-agnostic setting.} All methods are benchmarked on a standard H200 environment \textit{without} specialized quantization or sparsity kernels. Under this setting, Wanda 2:4 pruning drops to 42\% (LLM-only) or 28\% (full-model) SR with no speedup or memory benefit, and INT4 quantization preserves 94\% SR and saves memory but is \textit{slower} (0.61$\times$ action, 0.55$\times$ task) due to dequantization overhead. These results reflect the absence of optimized runtime support; with dedicated low-bit or structured-sparsity kernels, quantization and pruning methods can achieve substantially better latency.

\textit{Attention sublayers are highly redundant; MLP sublayers are not.}
Zero-shot CosSim-based attention dropping removes up to 8 attention sublayers (25\% of all attention) with no SR loss (Attn Drop 8: 98\%, task speedup 1.09$\times$), confirming large attention redundancy. In stark contrast, removing even 4 MLP sublayers causes complete failure (0\% SR, task speedup 0.42$\times$). The MLP BI scores have 10$\times$ lower variance than attention BI scores, yet the cosine-similarity metric fundamentally underestimates MLP criticality. This asymmetry explains why zero-shot \textit{block} dropping (which removes both attention and MLP) performs poorly (Block Drop 4: 78\%, task speedup 0.72$\times$), and why DTR's fine-tuning phase is essential to recover from MLP removal.

\section{Edge acceleration requires hardware-kernel alignment}
\label{app:edge_acceleration}

A practical motivation for \DTR is that reducing parameter count or theoretical FLOPs does not necessarily produce proportional wall-clock speedups on edge robotic platforms. In deployment, a compressed model must match both the hardware capabilities of the target device and the optimized kernels provided by the inference runtime. This is especially important for closed-loop robotic control, where latency affects control bandwidth and end-to-end task completion time.

\paragraph{Low-bit quantization.}
Low-bit quantization can reduce memory bandwidth and increase arithmetic throughput, but only when the target runtime supports the corresponding quantized operators. For example, Jetson Thor exposes Blackwell-class low-precision compute, including FP4 capability, while TensorRT supports quantized formats such as INT8, INT4, FP8, and FP4~\citep{nvidia2026jetsonthor,nvidia2026tensorrt_quantized}. However, a GPTQ- or INT4-compressed model does not automatically become faster on an edge device: the relevant linear, attention, and normalization operators must be lowered into efficient low-bit kernels. Otherwise, dequantization overhead, unsupported operators, or fallback execution can reduce or eliminate the expected speedup.

\paragraph{Sparse pruning.}
Sparsity-based pruning has an analogous constraint. Hardware acceleration for sparsity typically assumes a specific structured pattern rather than arbitrary zeros. NVIDIA sparse Tensor Cores, for instance, target fine-grained 2:4 structured sparsity, where two values in each group of four are zero~\citep{nvidia2023ampere_sparsity}. TensorRT also requires sparse weights to satisfy the supported structure and to be executed under compatible precision modes~\citep{nvidia2026tensorrt_sparsity}. As a result, generic unstructured pruning, or sparse weights that do not match the required pattern, may reduce model size without yielding reliable latency gains on Jetson-class devices.

\paragraph{Implication for DTR.}
DTR avoids this hardware-kernel alignment issue by physically removing transformer blocks and producing a smaller standard dense model. The resulting network uses ordinary dense operators with fewer layers, so its speedup does not rely on low-bit arithmetic, sparse Tensor Cores, or specialized sparse kernels. This is why we treat hardware friendliness separately from theoretical FLOPs reduction: quantization and sparsity can be effective when the deployment stack supports them, but DTR provides a simpler and more portable path to acceleration on edge robotic systems.

\end{document}